\documentclass{ieeeaccess}
\usepackage{cite}
\usepackage{amsmath,amssymb,amsfonts}
\usepackage{algorithmic}
\usepackage{graphicx}
\usepackage{textcomp}
\usepackage{caption}
\usepackage{acronym} 
\usepackage{multirow}
\usepackage{rotating}
\usepackage{rotating}
\usepackage{hyperref}
\usepackage{booktabs} 
\usepackage[tight,footnotesize]{subfigure}

\newcommand{\kk}[1]{}
\usepackage[none]{hyphenat}
\def\BibTeX{{\rm B\kern-.05em{\sc i\kern-.025em b}\kern-.08em
    T\kern-.1667em\lower.7ex\hbox{E}\kern-.125emX}}
\DeclareUnicodeCharacter{2212}{-}

\begin{document}
\history{Date of publication xxxx 00, 0000, date of current version xxxx 00, 0000.}
\doi{10.1109/ACCESS.2017.DOI}

\title{Navigating Threats: A Survey of Physical Adversarial Attacks on LiDAR Perception Systems in Autonomous Vehicles}





\author{\uppercase{Amira Guesmi}, and
\uppercase{ Muhammad Shafique}.
}
\address{eBrain Lab, Division of Engineering, New York University (NYU) Abu Dhabi, UAE}

\tfootnote{
This work was partially funded by the NYUAD Center for Cyber Security (CCS), funded by Tamkeen under the NYUAD Research Institute Award G1104 and the NYUAD Center for Interacting Urban Networks (CITIES), funded by Tamkeen under the NYUAD Research Institute Award CG001.
}

\markboth
{Guesmi \headeretal: Navigating Threats: A Survey of Physical Adversarial Attacks on LiDAR Perception Systems in Autonomous Vehicles
}
{Guesmi \headeretal: Navigating Threats: A Survey of Physical Adversarial Attacks on LiDAR Perception Systems in Autonomous Vehicles
}

\corresp{Corresponding author: Amira Guesmi (e-mail: ag9321@nyu.edu).}

\begin{abstract}

Autonomous vehicles (AVs) rely heavily on LiDAR (Light Detection and Ranging) systems for accurate perception and navigation, providing high-resolution 3D environmental data that is crucial for object detection and classification. However, LiDAR systems are vulnerable to adversarial attacks, which pose significant challenges to the safety and robustness of AVs. This survey presents a thorough review of the current research landscape on physical adversarial attacks targeting LiDAR-based perception systems, covering both single-modality and multi-modality contexts. We categorize and analyze various attack types, including spoofing and physical adversarial object attacks, detailing their methodologies, impacts, and potential real-world implications. Through detailed case studies and analyses, we identify critical challenges and highlight gaps in existing attacks for LiDAR-based systems. Additionally, we propose future research directions to enhance the security and resilience of these systems, ultimately contributing to the safer deployment of autonomous vehicles.
\end{abstract}

\begin{keywords}

Autonomous Vehicles, LiDAR, Adversarial Attacks, Perception Systems, Autonomous Driving, Spoofing Attacks, Physical Adversarial Objects, Reflective Objects, Sensor Fusion, Robustness, Anomaly Detection, Machine Learning, Signal Processing, Safety, Security, 2D Object Detection, Sensor Manipulation, Navigation, Real-world Scenarios, Defense Mechanisms, Point Cloud, 3D Object Detection
\end{keywords}

\titlepgskip=-15pt

\maketitle
\section{Introduction}



In recent years, there has been a surge in research focused on automated driving, with the first autonomous vehicles already in use. For instance, in California, Google subsidiary Waymo has launched its self-driving vehicles in various cities \cite{Waymo}, and in Beijing, Baidu Inc. has introduced the first autonomous cabs \cite{Baidu}. Additionally, drivers of non-autonomous cars benefit from an increasing number of assistance systems designed to enhance safety and ease of driving, paving the way toward fully autonomous cars \cite{jimenez2016advanced}. SAE International has defined stages (or levels) of automated driving \cite{levels}, with Level 5 representing full automation, where the vehicle can handle every possible situation autonomously, without any driver intervention. 

\begin{figure}
    \centering
    \includegraphics[width=\linewidth]{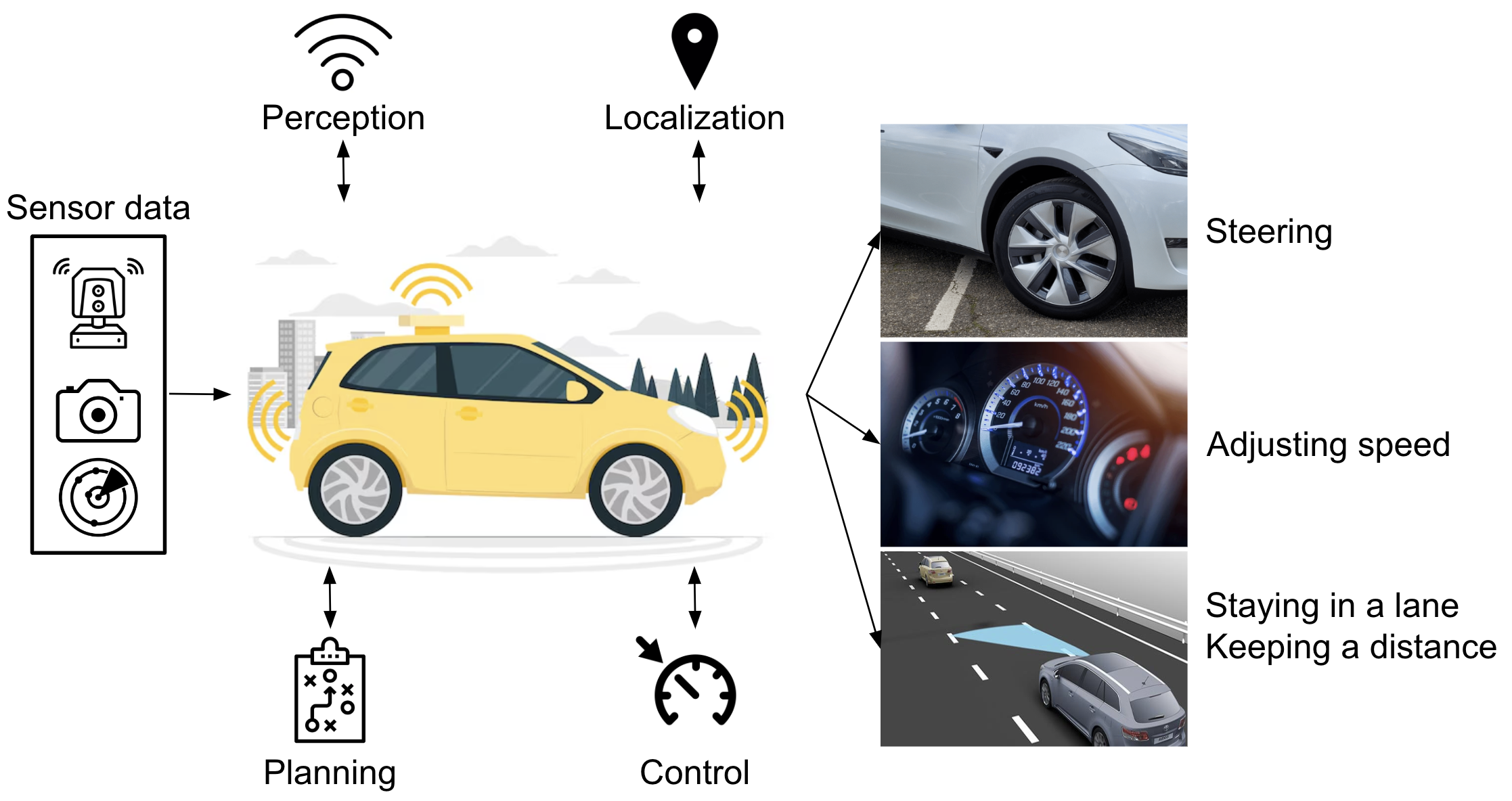}
    \caption{Overview of an autonomous vehicle navigation system: key modules including perception, localization, planning, and control.}
    \label{fig:car}
\end{figure}

\begin{figure}
    \centering
    \includegraphics[width=\linewidth]{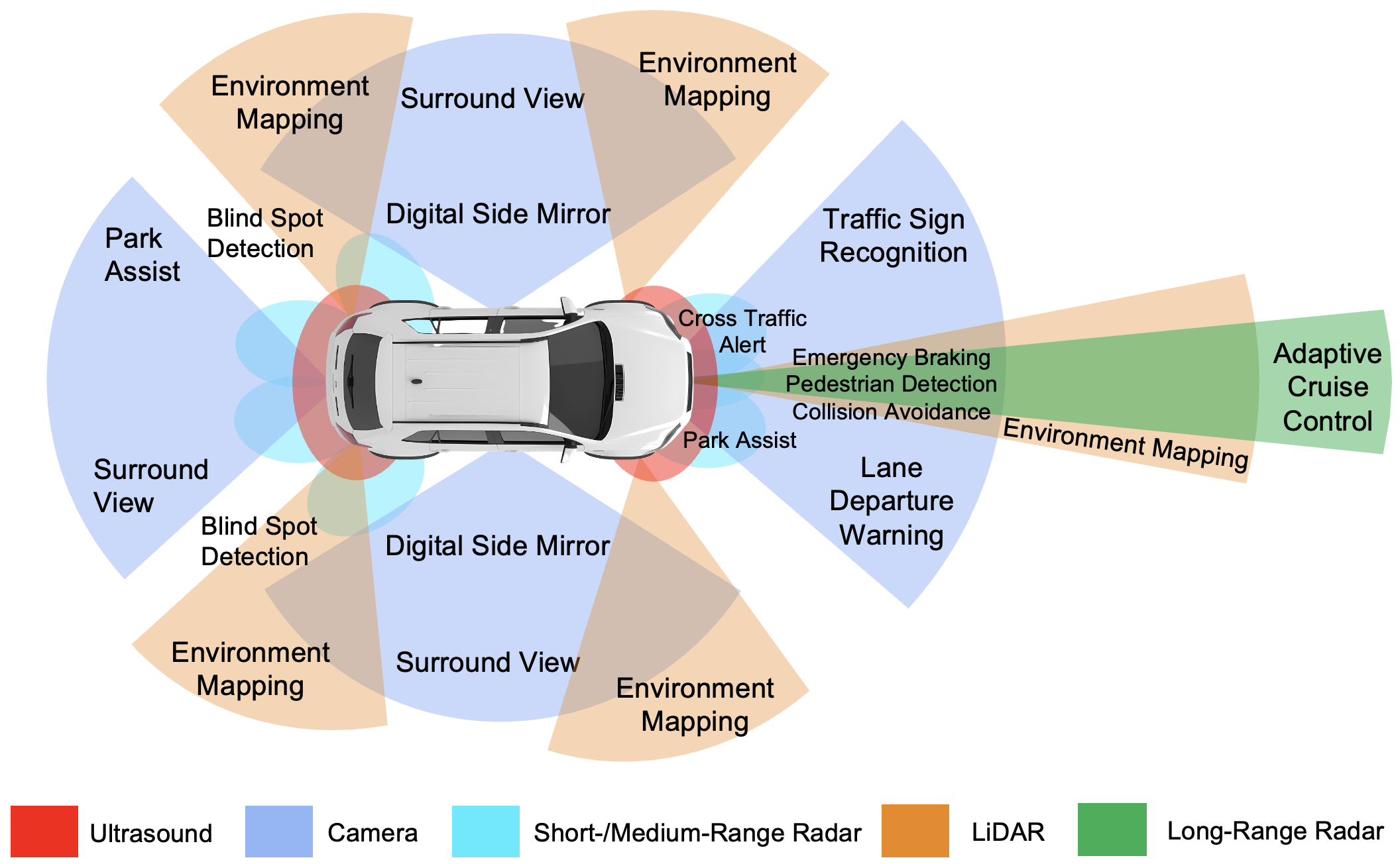}
    \caption{Perception sensors employed in autonomous vehicles.}
    \label{fig:sensors}
\end{figure}

Autonomous driving (AD) systems typically involves breaking down the complex task of autonomous driving into four primary interconnected modules (Figure \ref{fig:car}):

1. \textit{Perception}: This module focuses on interpreting the vehicle's environment using a combination of sensors like cameras, LiDAR, RADAR, and ultrasonic sensors (Figure \ref{fig:sensors}). It gathers information about the vehicle's surroundings, identifying objects, traffic signs, lane markings, and other elements in real time. The goal is to accurately detect and classify obstacles, pedestrians, and other vehicles to ensure safe navigation.

2. \textit{Localization}: This module determines the vehicle's exact position within a map or predefined route. GPS, inertial measurement units (IMUs), and sometimes visual odometry help the vehicle understand where it is, ensuring accurate positioning, which is critical for decision-making and route planning. High-definition maps often complement sensor data for precise localization.

3. \textit{Planning}: The planning module interprets the data from perception and localization to decide on the vehicle's next actions. This can include path planning, which generates a safe and efficient route, and behavior planning, which focuses on short-term actions like lane changes, stops, and adjusting speed according to traffic rules, road conditions, and safety margins.

4. \textit{Control}: This module converts the planned route and behavior into actual driving actions, managing the steering, throttle, braking, and speed control of the vehicle. The control system ensures that the vehicle accurately follows the planned trajectory while maintaining safety and comfort.

These modules work in tandem to enable the vehicle to drive autonomously, with constant feedback between perception, planning, and control. Each module plays a critical role in ensuring that the vehicle can navigate complex environments safely and efficiently.

Higher levels of automation require extensive information about the vehicle's surroundings, necessitating the development of advanced sensors to improve perception capabilities. One such sensor is the LiDAR system, which measures the distance to objects using the time of flight of emitted light beams \cite{bastos2021overview}. Unlike camera-based object detection, LiDAR does not rely on perfect visibility conditions, making it effective both day and night \cite{wallace2020full}. Complete environmental coverage is often achieved by deploying a 360° laser scanner mounted on the vehicle's roof, or by installing multiple stationary LiDAR sensors and combining their data. For example, the Audi Autonomous Driving Dataset (A2D2) used five LiDAR sensors mounted on a vehicle's roof \cite{geyer2020a2d2}. It is anticipated that future vehicles with autonomous driving functions will incorporate LiDAR systems along with other sensor technologies to ensure accurate three-dimensional object detection and safe automated driving \cite{sun2020}.

Machine learning algorithms, particularly deep neural networks, are increasingly used for recognizing and classifying 3D objects and 2D images. These algorithms aim to recognize patterns (objects) in the data and classify these objects into known categories such as cars, pedestrians, or bicycles. However, the decision-making process of these algorithms is often opaque, making it challenging to understand how specific recognition or classification outcomes are reached.

Given its crucial role in autonomous vehicle control and its reliance on external data, the LiDAR system is a potential target for adversarial attacks. These attacks intentionally introduce changes to the input data to induce detection/classification errors, potentially leading to erroneous warnings or accidents, with consequences ranging from property damage to loss of life \cite{petit2014potential, cao2019ccs, sun2020}. Therefore, it is imperative that detection models not only achieve high accuracy in object detection and classification but also exhibit resilience to adversarial attacks.


While there has been significant research into adversarial attacks in the 2D domain, such as attacks on image-based object detectors, the field of 3D object detection remains less explored. This is particularly true for physical attacks, which pose unique challenges due to the need for real-world implementation that accounts for sensor-specific constraints.
There are relatively few comprehensive surveys that focus specifically on attacks targeting 3D object detection systems, and even fewer that provide an in-depth exploration of physical attacks. For instance, \cite{li2024survey} presents a survey on attacks against both 2D and 3D object detection systems but covers only few examples of physical attacks in limited detail.
Similarly, \cite{mahima2024survey} briefly discusses 12 physical attacks, focusing primarily on spoofing and adversarial object-based techniques. In contrast, our survey provides an in-depth analysis of 19 physical attacks, offering detailed insights into a broader range of techniques.


In this survey, we focus on the specific vulnerabilities of LiDAR systems to adversarial attacks, highlighting the various methods used by attackers to deceive or disrupt the perception capabilities of these sensors. It examines different types of attacks, such as spoofing, using reflective objects, and designing adversarial objects, and their impacts on the accuracy and reliability of LiDAR-based perception systems in autonomous vehicles.
The primary goals of this survey are to consolidate existing knowledge on adversarial attacks related to LiDAR systems in autonomous vehicles, identify gaps in current research, and propose directions for future work. By providing a comprehensive review of the state-of-the-art techniques and their effectiveness, this survey contributes to the development of more secure and reliable LiDAR-based perception systems, ultimately enhancing the safety and performance of autonomous vehicles.

\textbf{Our contributions are as follows:}

\begin{itemize}
    \item We propose, to the best of our knowledge, the first comprehensive taxonomy of physical adversarial attacks specifically targeting LiDAR-based systems for 3D object detection and provide a detailed characterization of existing attack methods.
    \item We discuss the various challenges and physical constraints associated with these attacks, highlighting the complexities involved in both executing and defending against them.
    \item We review the evaluation metrics, datasets, models, simulators, and autonomous driving platforms utilized in current research to assess the performance and robustness of perception systems under adversarial attacks.
    \item We identify open research areas and suggest potential future directions. 
    
\end{itemize}

The structure of the remaining article is organized as follows. Section \ref{background} provides an overview of the preliminaries, including essential topics and concepts related to LiDAR technology and adversarial attacks. Section \ref{taxonomy} delves into the taxonomy of adversarial attacks against LiDAR systems, discussing various forms such as physical spoofing, reflective objects, and adversarial physical objects. Section \ref{metrics} examines attack capabilities and scenarios, exploring how attackers exploit system-level and environmental-level knowledge to manipulate LiDAR perception. Section \ref{techniques}, discuss different attack design challenges and physical constraints.
In Sections \ref{lidar_perception} and \ref{fusion_perception}, we review the evaluation metrics and tools used to assess the effectiveness of adversarial attacks and the robustness of perception systems for both LiDAR-based and sensor fusion-based, respectively. 
Section \ref{defenses} discusses defense mechanisms designed to counteract these attacks, covering both model-agnostic and model-based approaches. Section \ref{discussion} highlights open research challenges and suggests future research directions, emphasizing the need for improved defenses and robustness in real-world environments. Finally, Section \ref{conclusion} concludes the article with a summary of the key insights and contributions, underscoring the importance of enhancing the security and reliability of LiDAR-based perception systems in autonomous vehicles.

\section{Background}
\label{background}
\subsection{Deep Learning Based Perception in AV Systems }
In Autonomous Vehicles, one fundamental pillar is perception, which leverages sensors like cameras, LiDARs, and radars to understand the driving environment.

\subsubsection{Camera-Based Perception.} 
The advancements in deep learning for object detection and instance segmentation have driven the integration of these models into perception modules of autonomous vehicle systems. Camera-based solutions use visual images to identify the locations, types, and orientations of objects on the road \cite{wang2019pseudo}. However, these methods face inherent challenges, particularly in estimating depth from 2D images, which leads to suboptimal performance in 3D localization tasks \cite{guesmi2024saam, guesmi2023aparate, guesmi2024ssap}. This limitation makes camera-based perception systems vulnerable to errors in accurately detecting the spatial position of objects in the real world.

Moreover, camera-based systems have become prime targets for various adversarial attacks in real-world environments. These attacks exploit the system's sensitivity to visual perturbations, including physical adversarial examples, such as manipulated stickers or patches, that can cause incorrect detection or misclassification \cite{guesmi2023physical, guesmi2024dap, guesmi2023advart}. The susceptibility of these systems to adversarial attacks underscores the need for robust defense mechanisms, especially given the critical role of camera-based perception in ensuring the safety of AV operations.

\subsubsection{LiDAR-Based Perception.} 
LiDAR sensors function by emitting laser pulses and capturing their reflections using photodiodes. Since the speed of light is constant, the time taken for the reflected pulse to return to the sensor provides an accurate measurement of the distance between the LiDAR and surrounding objects. By firing laser pulses at various vertical and horizontal angles, LiDAR generates a 3D point cloud that AV systems use to detect and localize objects.

LiDAR is an active sensor capable of producing high-resolution 3D point cloud data, making it crucial for accurately mapping the surrounding environment. The most common LiDAR technology today is direct time-of-flight (dToF) LiDAR, which measures the time difference between the emission and reflection of laser pulses to determine the precise 3D coordinates of points on object surfaces. By sweeping laser pulses across horizontal (azimuth) and vertical (elevation) angles, the system constructs a detailed 3D representation of the scene, known as a point cloud. These systems often operate with high peak laser power (e.g., 10W), allowing LiDAR to detect objects over 200 meters away, even in challenging conditions such as bright sunlight or low ambient light.

\noindent\textbf{First-Generation LiDARs:}
Early models like the VLP-16 \cite{vlp16} and VLP-32c \cite{vlp32c} were pioneering in the development of LiDAR for autonomous vehicles. They provided relatively low-density point clouds and shorter detection ranges but laid the foundation for future advancements in 3D perception.

\noindent\textbf{Next-Generation LiDARs:} 
Next-generation LiDARs, such as the VLS-128 \cite{vls128}, Pixell \cite{pixell}, OS1-32 \cite{os1}, and Realsense L515 \cite{realsense}, offer significantly higher resolution, better accuracy, and enhanced range. These systems generate denser point clouds, enabling more detailed object detection and improved performance in complex environments. Other advancements include the Horizon \cite{horizon}, XT32 \cite{xt32}, and Helios 5515 \cite{helios}, which further enhance scanning capabilities, robustness, and integration for various autonomous vehicle platforms. These newer models reflect the trend towards more compact, powerful, and efficient LiDAR systems that are increasingly crucial for safe and reliable AV operations.

\begin{sidewaystable}
    \centering
    \small 
    \caption{Comparison of LiDAR technologies: 1st-G and New-G: First and new-generation. FOV: Field of View. (Table adapted from \cite{Sato2023})}
    \begin{tabular}{ccccccccccc}
         \hline
         & &  \multicolumn{3}{c}{Velodyne}  & Leddar & Ouster & Intel & Livox & Hesai  &  Robosense  \\
          \cline{3-9}
         &     & VLP-16 \cite{vlp16} & VLP-32c \cite{vlp32c} & VLS-128 \cite{vls128}& Pixell \cite{pixell}& OS1-32 \cite{os1}& Realsense L515 \cite{realsense}& Horizon \cite{horizon}& XT32 \cite{xt32}& Helios 5515 \cite{helios}\\
         \hline
         \parbox[t]{2mm}{\multirow{8}{*}{\rotatebox[origin=c]{90}{General Specs}}} & Gen. (year) &1st-G (2016)  & 1st-G (2017) & 1st-G (2017) & New-G (2019) & New-G (2019) & New-G (2019) & New-G (2020)  &  New-G (2020) &  New-G (2021)  \\
         & Scanning Type & Rotating & Rotating & Rotating & Flash & Rotating & MEMS & MEMS & Rotating  &  Rotating  \\
         & Wavelength & 905 nm  & 905 nm & 905 nm & 905 nm & 865 nm & 860 nm & 905 nm & 905 nm  &  905 nm  \\
         & Vertical FOV & $30^\circ$ & $40^\circ$ & $40^\circ$ & $16^\circ$ & $45^\circ$ & $55^\circ$ & $25.1^\circ$ & $31^\circ$  &  $70^\circ$  \\
         & Horizontal FOV & $360^\circ$ & $360^\circ$ & $360^\circ$ & $180^\circ$ & $360^\circ$ & $70^\circ$ & $81.7^\circ$ & $360^\circ$  & $360^\circ$   \\
         & Max. Range [m] & 100 & 200 & 300 & 56 & 120 & 9 & 260 & 120  & 150   \\
         & Min. Range [m]  & 1 & 1 & 0.5 & 0.1 & 0.3 & 0.25 & 0.5 & 0  & 0.2   \\
         & Vertical Channel  & 16 & 32 & 128 & 8 & 32 & - & - & 32  &  32  \\
         \hline
        \parbox[t]{2mm}{\multirow{3}{*}{\rotatebox[origin=c]{90}{Security}}} & Simul. Firing & 1 & 2 & 8 & 3 & 32 & 1 & 1 & 1  &  1  \\
         & Timing Random &  &  &  & \checkmark & \checkmark & \checkmark & \checkmark &   &  \checkmark  \\
         & Finger printing &  &  &  &  &  &  &  & \checkmark  &    \\
         \hline
    \end{tabular}
    \label{tab:lidar}
\end{sidewaystable}

Table \ref{tab:lidar} provides a comparison of various LiDAR technologies, distinguishing between first-generation (1st-G) and new-generation (New-G) systems. It highlights key advancements in technology, such as scanning types, field of view (FOV), and range, as well as improvements in security features, including timing randomness and fingerprinting.
\subsubsection{Multi-Sensor Fusion Based Perception.} 

In high-level autonomous driving systems, particularly those at Level 4 and above \cite{levels}, perception is a critical module responsible for real-time detection of surrounding objects. Given its direct impact on safety-critical decisions, such as collision avoidance and path planning, advanced AD systems (e.g., Google Waymo, Pony.ai, Baidu Apollo) predominantly adopt Multi-Sensor Fusion (MSF)-based designs to enhance the accuracy and robustness of perception \cite{Waymo, Baidu, autoware, Pony_ai, nvidia}.

\noindent\textbf{MSF Design Principle and Assumptions:}
In MSF-based perception, the system fuses data from multiple sensor modalities—typically cameras and LiDAR—to leverage the strengths of each and compensate for their individual limitations. Cameras, for instance, excel at capturing high-resolution texture and color information but struggle with depth estimation \cite{frossard2018end}. LiDAR, on the other hand, provides accurate 3D spatial data but lacks the ability to capture color or texture \cite{liang2018deep}. By combining these inputs, MSF algorithms aim to improve object detection performance, yielding higher accuracy and robustness than either modality could achieve independently \cite{chen2017multi, xu2018pointfusion, liang2019multi}. The core assumption of MSF is that, under most conditions, at least one sensor will provide reliable information, allowing the system to maintain accurate perception in complex environments.

\begin{figure}
    \centering
    \includegraphics[width=\linewidth]{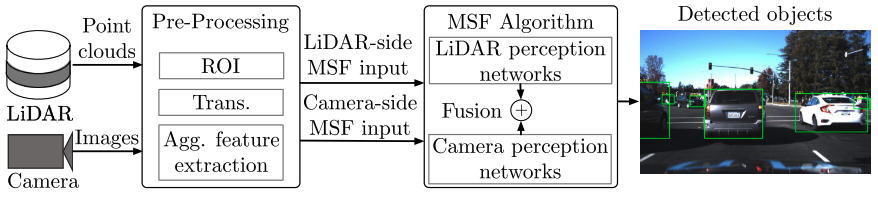}
    \caption{Overview of multi-sensor fusion based AD perception design. (Figure adapted from \cite{cao2021}).}
    \label{fig:fusion}
\end{figure}

\noindent\textbf{Representative MSF Algorithm Designs:}
State-of-the-art MSF algorithms in AD perception predominantly fuse data from two key sources: cameras and LiDAR \cite{frossard2018end, chen2017multi, liang2018deep, ku2018joint, du2018general}. Figure \ref{fig:fusion} illustrates a typical MSF-based AD perception design. Before fusion, raw inputs from cameras (images) and LiDAR (point clouds) are pre-processed to prepare them for the MSF algorithm \cite{Baidu, autoware}. Pre-processing steps often include data transformations (e.g., rotations and translations), Region of Interest (ROI) filtering to remove irrelevant data, and feature extraction to aggregate useful information. These steps reduce the size and complexity of the input data, improving the efficiency and performance of the MSF algorithm at runtime \cite{bello2020deep}.

MSF designs typically use deep neural networks (DNNs) to process the camera and LiDAR inputs independently, known as the camera perception network and the LiDAR perception network, respectively \cite{Waymo, Baidu, autoware}. The results from these two networks are then fused using either (1) DNN-based fusion techniques \cite{liang2019multi, du2017car}, which can occur at an early or late stage in the processing pipeline, or (2) rule-based fusion methods \cite{Baidu, autoware}, where pre-defined rules are used to combine the outputs at the final stage. DNN-based fusion allows for deeper integration of the data, often leading to higher accuracy, while rule-based fusion offers modularity, easier interpretability, and flexibility for integrating different models \cite{yurtsever2020survey, chi2017deep}.

\noindent\textbf{Data Pre-Processing and Aggregation:}
Pre-processing steps, especially for LiDAR, are crucial given the large volume of data generated. A single LiDAR sensor can capture millions of 3D points per second \cite{Velodyne}, so processing raw point clouds in real-time can be computationally expensive. To address this, many LiDAR-based perception models use aggregation techniques that summarize the point cloud into more manageable 3D cells or voxels, which encode key features such as average height, intensity, and point density \cite{engelcke2017vote3deep, zhou2018voxelnet, lang2019pointpillars, wang2015voting}. Some designs even convert 3D point clouds into 2D Bird’s-Eye View (BEV) representations to further enhance real-time performance \cite{yang2018pixor}. BEV has become a widely adopted approach in industry-grade AD systems \cite{Baidu, chen2017multi, beltran2018birdnet}, as it simplifies 3D data into a 2D plane while retaining critical spatial information.

\subsection{Adversarial Attack for 2D Image}

Szegedy et al. \cite{szegedy2013intriguing} first uncovered adversarial examples, where subtle, carefully crafted perturbations are added to 2D images. These perturbations are often imperceptible to the human eye, yet they can cause deep learning models to misclassify the input. Since this discovery, numerous works have further explored and developed more complex and effective adversarial attacks across various domains \cite{carlini2017towards, goodfellow2014explaining, moosavi2016deepfool, papernot2016limitations, guesmi2023advrain}. These adversarial attacks highlight critical vulnerabilities in deep learning models, particularly in computer vision.

In response, several defense mechanisms have been proposed to mitigate the impact of adversarial examples. One widely studied approach is adversarial training, as introduced by Tramer et al. \cite{tramer2017ensemble}, where models are trained with known adversarial examples to improve their robustness. Various extensions of adversarial training have since been explored to further enhance model resilience \cite{guesmi2022room, guesmi2024exploring}.

Another defense mechanism is defensive distillation \cite{papernot2016distillation}, which retrains the model to smooth adversarial gradients. This process distills knowledge from the original training phase and forces the adversarial output vectors to converge, making it harder for attackers to deceive the model. Randomized pre-processing techniques, as proposed by Guesmi \cite{guesmi_sit}, can also be used to filter out potential adversarial inputs before they reach the model.

In addition, dimensionality reduction-based defenses have gained traction, as they aim to neutralize adversarial noise by reducing the complexity of the input space. Techniques proposed by \cite{chattopadhyay2023defensivedr, chattopadhyay2024anomaly, chattopadhyay2023oddr} leverage this approach to diminish the impact of adversarial perturbations, providing an additional layer of defense against attacks.

\section{TAXONOMY OF ADVERSARIAL ATTACKS}
\label{taxonomy}
\begin{sidewaysfigure}
    \centering
    \includegraphics[width=1\linewidth]{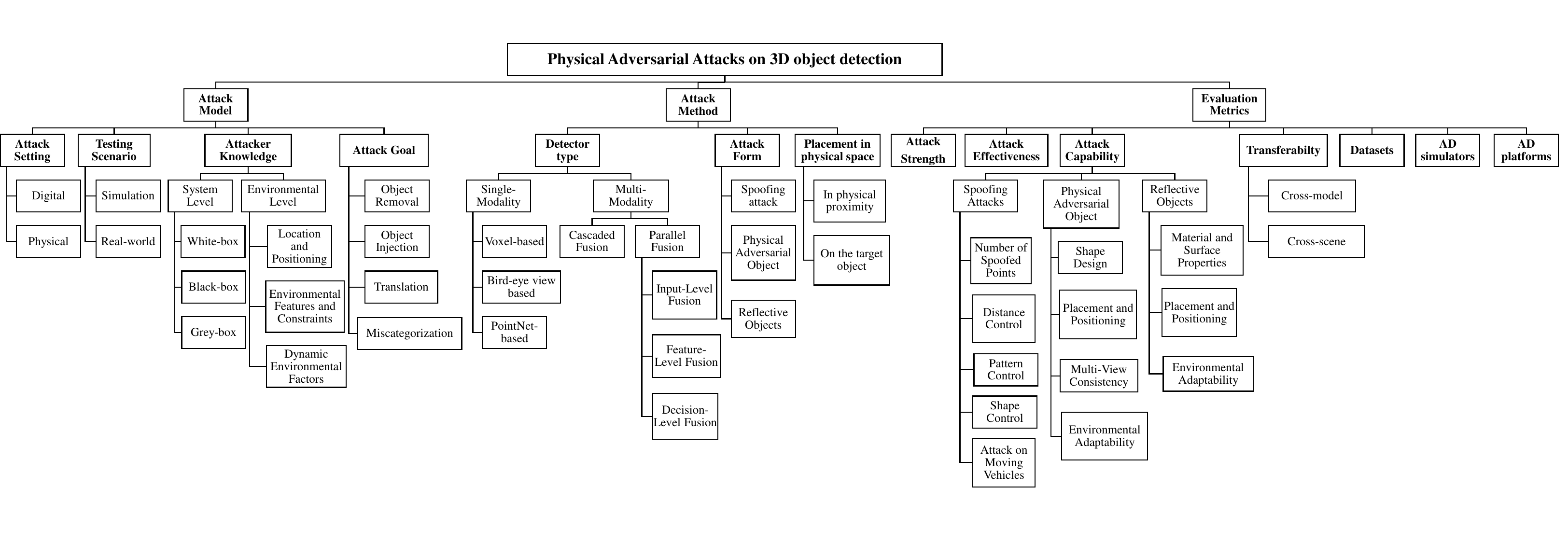}
    \caption{Taxonomy and evaluation metrics for adversarial attacks on LiDAR-based perception systems used in 3D object detection.}
    \label{fig:taxonomy}
\end{sidewaysfigure}

We present our taxonomy of adversarial attacks for object detection in Figure \ref{fig:taxonomy}. We describe preliminary information in Section \ref{preliminaries}, followed by detailed categorizations for attack model and attack method in Section \ref{model} and \ref{method}, respectively.

\subsection{Preliminaries}
\label{preliminaries}
Given the input scene point cloud $S$ and the ground truth location $G$ of the target objects 
, the point cloud $X= \{x_i\}^{N'}_{i=1}$ of a target can be obtained, satisfying $X \in S$. We apply point-wise perturbations to the target point cloud $X$, and generate adversarial point clouds to attack a deep $3D$ object detection model.

Denote $\{\delta_i\}^{N'}_{i=1}$ as the perturbation vectors, satisfying $\delta_i \in \mathbb{R}^3$, added to the point cloud of the target $X$, producing adversarial point cloud $X'= \{x_i + \delta_i\}^{N'}_{i=1}$. A target proposal $p_i$ generated by a deep $3D$ object detection model can be called positive, if its Intersection of Union (IoU) with ground truth bounding box of a corresponding object is larger than a threshold $\mu$ and at the same time its confidence score is greater than a threshold $\alpha$. The others are defines as negative. 

The adversarial attack can be formulated as an unconstrained optimization problem \cite{cai2020}, and the loss function are defined as follows:

\begin{equation}
    L = L_{adv} + \lambda L_{per},
\end{equation}

where $L_{adv}$ and $L_{per}$ stand for the corresponding adversarial loss and point cloud perturbation loss, and $\lambda$ is used as a weighing parameter.

The adversarial loss $L_{adv}$ is defined on the whole set of positive proposals, and is designed to minimize both their Intersection of Union with the ground truth and the associated confidence, as follows:

\begin{equation}
    L_{adv} = \sum_{(p_i, s_i)\in P} -IoU(p_i,p^*)log(1-s_i),
\end{equation}

where $P$ represents the whole set of positive target proposals with $p_i$, $s_i$ standing for the $i$th bounding box and the associated confidence. $p^*$ is denoted as the ground truth of the bounding box. The operator $IoU(\dot,\dot)$ calculates the intersection between two bounding boxes.

To make sure that the generated adversarial point cloud is visually imperceptible to humans, the perturbation loss $L_{per}$ is added. A number of distance metrics have been proposed to measure the deviation of the adversarial point clouds from the origin, such as Hausdorff distance, chamfer distance. In this paper, we choose the $L_2$ norm distance for its efficiency both in performance and computation. Thus given the target point cloud $X$, the perturbation vectors $\{\delta_i\}^{N'}_{i=1}$, and the adversarial point cloud $X'= \{x_i + \delta_i\}^{N'}_{i=1}$, the perturbation loss $L_{per}$ can be defined as follows:

\begin{equation}
    L_{per} = \sum^{N'}_{i=1}\parallel\delta_i\parallel^2.
\end{equation}

To sum up, the loss function defined for the adversarial attack is defined as follows:

\begin{equation}
    L_{adv} = \sum_{(p_i, s_i)\in P} -IoU(p_i,p^*)log(1-s_i) + \lambda\sum^{N'}_{i=1}\parallel\delta_i\parallel^2.
\end{equation}

Given the loss function defined on the point cloud perturbations, the adversarial attack can be realized by the means of differentiation. To be specific, the optimization process is guided by the gradient computed from the loss function with respect to each point. $\lambda$ defined in the loss function is set as a constant $0.1$.

\subsection{Attack Model}
\label{model}

\subsubsection{Attack Setting} 

\textbf{Digital}: Digital attacks involve altering the points of point clouds input into the target model without interacting with the physical environment. In this type of attack, the attacker manipulates the data directly within the digital realm, meaning they do not interfere with the actual physical objects or the environment being scanned. The attacker also lacks access to the processes of information capture, pre-processing, and post-processing, focusing solely on the digital manipulation of the input data to deceive the target model.

\begin{figure*}
    \centering
    \includegraphics[width=0.8\linewidth]{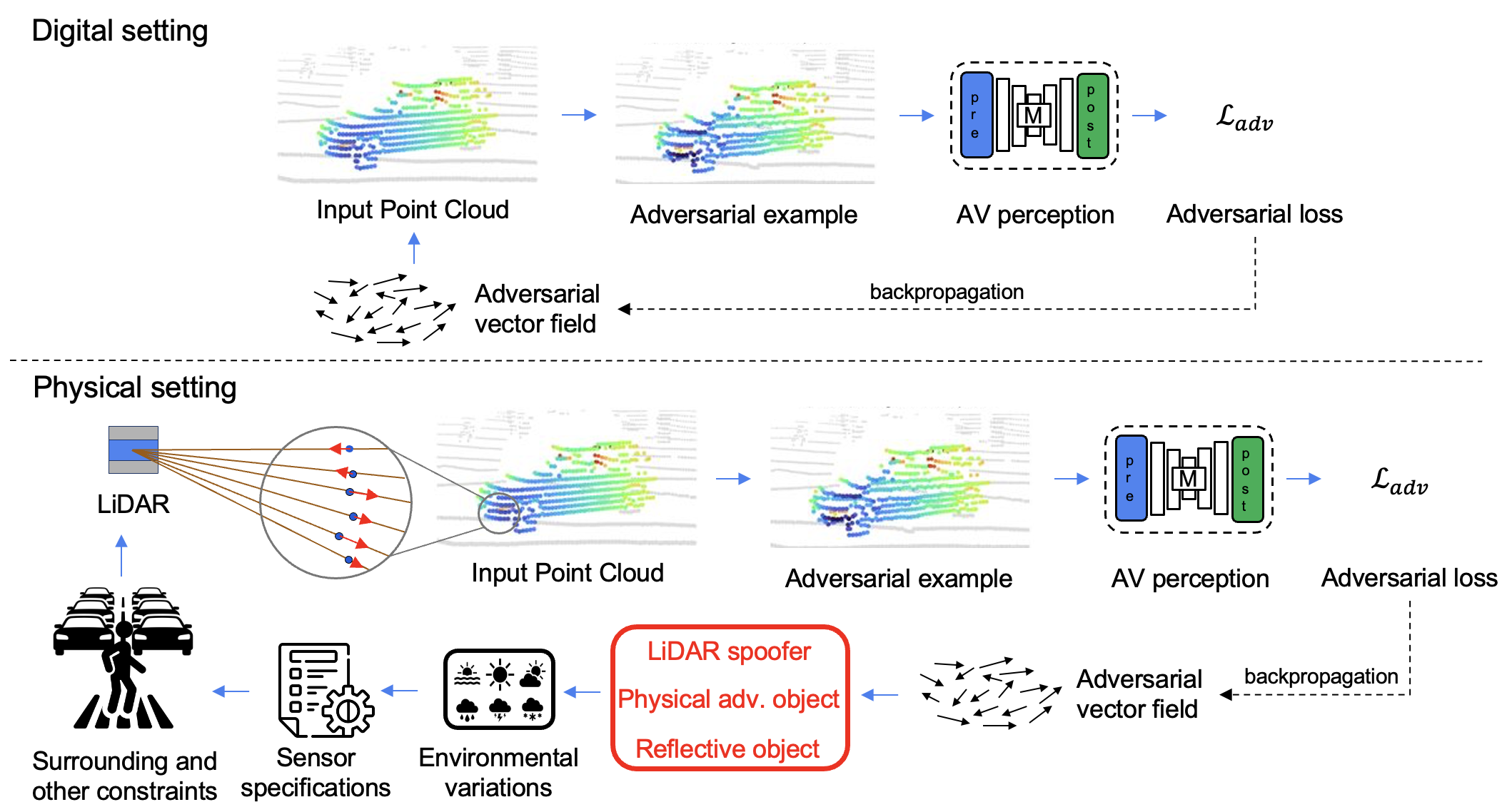}
    \caption{Comparison of Digital and Physical Attacks.}
    \label{fig:physical}
\end{figure*}

\noindent\textbf{Physical}: Physical attacks, on the other hand, involve direct manipulation of the physical environment or objects within the LiDAR sensor's field of view. In this scenario, the attacker alters real-world objects or settings in ways that affect how they are perceived by the LiDAR system. This could include placing reflective materials, altering the shape or surface of objects, or introducing new objects to the environment to create misleading or false data in the point cloud. Unlike digital attacks, physical attacks exploit the real-world conditions that the LiDAR sensor captures, aiming to disrupt the system’s ability to accurately detect and classify objects (see Figure \ref{fig:physical}).
We focus on physical attacks in this literature review as they present more realistic and practical threats to autonomous vehicle systems, highlighting the importance of addressing these vulnerabilities to ensure the safety and reliability of autonomous driving technologies.

\subsubsection{Test Scenario}

\textbf{Simulation}: Simulated attacks are performed within controlled, virtual environments where the entire system, including the LiDAR sensors and the physical environment, is modeled in software. This setting allows researchers and attackers to experiment with different attack techniques, pre-processing, post-processing, and synchronization challenges without the risks associated with real-world testing. Simulated environments enable precise control over the variables, making it easier to study the effects of various attack strategies and to refine them before potentially applying them in real-world scenarios. Common techniques in simulated attacks include generating synthetic point clouds that incorporate adversarial perturbations and testing the effects of these perturbations across different models and sensor configurations.

\noindent\textbf{Real-World}: Real-world attacks involve deploying adversarial techniques in actual environments where autonomous vehicles or LiDAR-equipped systems operate. These attacks are more challenging due to the complexities and unpredictability of real-world conditions, such as variable lighting, weather, and the dynamic nature of the environment. Real-world attacks require careful planning and execution, including synchronizing the timing of the attack with the vehicle’s sensors and considering the physical constraints of manipulating objects in a real environment. The attacker must also account for the entire sensor processing pipeline, from pre-processing (such as filtering noise from the point cloud) to post-processing (such as data fusion and decision-making algorithms), ensuring that the attack is effective across all stages. Real-world attacks are the most concerning as they pose direct threats to the safety and functionality of autonomous vehicles, making them a critical focus for defensive strategies.

\subsubsection{Attacker Knowledge}
\textbf{System Level.}\\
At the system level, Attacker Knowledge refers to the extent of information the attacker possesses about the target LiDAR-based perception system and its underlying algorithms. This knowledge significantly influences the design and potential success of the attack. It is typically categorized into three levels:

\noindent\textit{\textbf{White-box attacks}}: The attacker has complete knowledge of the target system, including the architecture of the perception model, parameter values (e.g., weights and biases of a neural network), training data, and the processing pipeline. This allows the attacker to craft highly targeted and effective adversarial attacks by exploiting specific vulnerabilities in the model. With this level of knowledge, the attacker can generate adversarial examples by calculating gradients directly from the model, ensuring a high success rate in fooling the perception system.

\noindent\textit{\textbf{Black-box attacks}}: The attacker has no direct access to the internal workings of the target system, including the model architecture, parameters, and training data. The attacker can only observe the system's input-output behavior, such as submitting input queries and observing the system's responses.
In this scenario, the attacker may rely on querying the system with different inputs and analyzing the outputs to craft adversarial examples. This often involves using surrogate models to approximate the target system and generate transferable attacks. Due to the lack of detailed system information, black-box attacks tend to be less precise and may require a higher number of queries to achieve the desired outcome.

\noindent\textit{\textbf{Grey-box attacks}}: The attacker has limited knowledge about the target system. This could include some information about the model architecture or a subset of the training data, but not the complete system. For example, the attacker might know the general type of model used (e.g., a convolutional neural network) but not the specific architecture or parameter values. With partial knowledge, the attacker can use this information to enhance the effectiveness of the attack compared to a black-box scenario. However, they still lack the full details needed for the high precision of white-box attacks. Grey-box attacks are generally more successful than black-box attacks but less so than white-box attacks, as the attacker can partially tailor the attack to the system's characteristics.


\noindent\textbf{Environmental Level.}\\ 
At the Environmental Level, Attacker Knowledge pertains to the attacker's understanding of the physical environment in which the autonomous vehicle operates. This knowledge includes the spatial and contextual information about the surroundings that can be exploited to design effective attacks on the LiDAR perception system. Key aspects of environmental-level attacker knowledge include:

\noindent\textit{\textbf{Location and Positioning}}: 
This includes knowledge about the approximate location of target objects, i.e., the attacker knows the approximate location of objects within the environment. This allows the attacker to strategically place adversarial elements, such as spoofed points or physical objects, in positions that can maximize their disruptive impact on the LiDAR system \cite{Kobayashi2024}.
Also, knowledge about the frustum region for spoof point placement. Understanding the frustum region (the cone-shaped area that the LiDAR scans) enables the attacker to insert spoof points within this region to deceive the perception system. Accurate placement within this region increases the chances of successful object injection or removal.

\noindent\textit{\textbf{Environmental Features and Constraints}}:
This includes knowledge of physical constraints, i.e., the attacker is aware of environmental constraints, such as terrain, occlusions, and reflective surfaces. This knowledge helps the attacker design attacks that either exploit these constraints (e.g., placing reflective objects to cause false readings) or avoid them (e.g., avoiding areas where objects might be occluded from the LiDAR's view).
Also, understanding of sensor limitations meaning being aware of the LiDAR's range, field of view, and sensitivity to environmental conditions (e.g., rain, fog) allows the attacker to craft attacks that work within these limitations or take advantage of them. For example, knowing the LiDAR’s vertical and horizontal angles helps in aligning spoof points with the sensor’s laser rays.

\noindent\textit{\textbf{Dynamic Environmental Factors}}:
The attacker understands how objects in the environment move relative to the vehicle, allowing for more sophisticated attacks that account for dynamic changes. This includes timing the placement of spoof points or physical objects to coincide with the vehicle's movement through the environment.
This can also include the knowledge of the behavior of other vehicles, pedestrians, and obstacles can help the attacker predict and manipulate the vehicle’s response to the altered perception data. For example, placing adversarial objects in areas where vehicles typically change lanes or stop can induce hazardous maneuvers.

\subsubsection{Attack Goal}

\textbf{Object Injection or false positive (FP)}: The goal of the adversary in object injection attacks is to cause the model to generate redundant bounding boxes or to detect and classify nonexistent objects within the input image. By injecting these false objects into the model's perception, the attacker aims to deceive the system into recognizing and responding to entities that are not actually present in the environment, potentially leading to erroneous decisions by the autonomous vehicle, such as unnecessary evasive maneuvers or incorrect navigation. 
The goal of achieving a false positive outcome is to force the victim to perform dangerous maneuvers (e.g., emergency braking or lane change) to avoid the false object. For example, LiDAR spoofing attacks can result in safety-critical incidents, as shown with Baidu’s Apollo \cite{Baidu, sun2020}.

\noindent\textbf{Object Removal or false negative (FN)}: The goal of the adversary in object removal attacks is to cause the model to miss detecting objects that are actually present in the environment. By manipulating the input data or the physical environment, the attacker aims to prevent the model from recognizing certain objects, effectively erasing them from the system’s perception. This can lead to dangerous situations, as the autonomous vehicle may fail to respond appropriately to obstacles, pedestrians, or other critical objects, increasing the risk of accidents.
The goal of achieving a false negative outcome is to remove an existing object from the perception output such that path planning and control are compromised. Such attacks can have the devastating consequence of the victim crashing into an unsuspecting object hidden to perception (e.g., as in \cite{cao2021}).

\noindent\textbf{Translation}: FP and FN outcomes are insufficient to fully capture the effects of perception attacks. Some cascaded semantic fusion architectures (e.g., FPN) enforce one-to-one matching between 2D and 3D detections; thus, an FP necessarily implies an FN. We call such instances translation outcomes as the attacker has created physical distance between the negated ground truth (FN) and the spoofed detection (FP). Translation outcomes may cause emergency braking if objects are moved to front-near positions or collision when moved farther from the victim or to a different lane.

\noindent\textbf{Miscategorization}: The goal of the adversary in miscategorization attacks is to perturb the input data in a way that causes the model to predict an incorrect label for a detected object.

\subsection{Attack method}
\label{method}

\subsubsection{Detector Type}
3D object detection plays an important role in the field of autonomous driving and has rapid development benefited from the breakthroughs in deep learning and sensor technologies.

\noindent\textbf{Single-Sensor Models}: Understanding the environment in 3D provides autonomous vehicles (AVs) with more comprehensive and reliable information compared to using 2D bounding boxes or pixel masks. With the growing adoption of 3D sensors, it's now possible to capture and process larger volumes of 3D data, delivering accurate depth information crucial for both mobile devices and AVs. Point clouds, which capture the 3D coordinates of points sampled from the surfaces of physical or virtual objects, are a widely used data format in 3D vision applications such as industrial modeling, surveying, and autonomous driving. Unlike images that consist of ordered pixels, point clouds are unordered, making them challenging to analyze using conventional deep learning techniques. To address these challenges, several methods have been developed to design deep learning models capable of processing point cloud data for tasks like classification and object detection. These methods can be broadly categorized into three types: Voxel-based approaches, Bird's-eye view techniques, and PointNet-based methods.

\noindent\textit{\textbf{Voxel-based Architecture:}} VoxelNet \cite{zhou2018voxelnet} organizes point clouds into uniformly spaced 3D voxels, enabling structured analysis. This method applies 3D Convolutional Neural Networks (CNNs) for predicting 3D bounding boxes, followed by a 2D convolutional detection layer in the final stage. Several recent studies \cite{lehner2019patch, kuang2020voxel, yan2018second} have adopted this voxel-based approach, achieving state-of-the-art results. PointPillar \cite{lang2019pointpillars} is another example of a voxel-based technique. It employs an encoder to transform features extracted from voxelized point clouds into sparse pseudo-images, with final predictions made by applying 2D CNNs and an SSD-based \cite{Liu_2016} detection network on these pseudo-images.

\noindent\textit{\textbf{Bird's-Eye View Architecture:}}  
Leveraging Deep Neural Networks to process 2D images is a well-established technique. Consequently, researchers have developed methods to convert LiDAR point clouds into ordered 2D structures for 3D object detection in AV systems. These approaches \cite{shi2020pv, yang2018pixor} transform the point cloud data into a bird's-eye view representation, which allows for more efficient processing using 2D convolutions. For instance, PV-RCNN \cite{shi2020pv}, known for its high performance in the KITTI bird's-eye view benchmark, is often used as a target model in black-box attack scenarios. Other notable examples of this approach include PIXOR \cite{yang2018pixor}.

\noindent\textit{\textbf{PointNet-Based Architecture:}}  
Bird's-eye view and voxel-based methods utilize deep learning models to analyze point cloud data, aiming to reduce computational costs and improve performance with sparse data. However, these methods cannot process raw point cloud data directly and instead require transformation techniques to convert the raw data into a more structured form that can be easily processed. While these transformations simplify the processing, they often lead to some degree of information loss, limiting the overall performance. In contrast, PointNet \cite{qi2017pointnet} and PointNet++ \cite{qi2017pointnet++} are designed to handle raw point cloud data directly by applying max-pooling and transformations to convert the unordered and dimensionally flexible input data into fixed-length global feature vectors. This approach enables end-to-end learning architectures on raw point cloud data, preserving more of the original information. PointNet is particularly robust due to its introduction of critical points and upper bounds concepts. PointNets are widely used in various applications, including 3D object detection \cite{shi2019pointrcnn, qi2018frustum}, where they serve as backbone networks for feature extraction.

\begin{figure}
    \centering
    \includegraphics[width=0.7\linewidth]{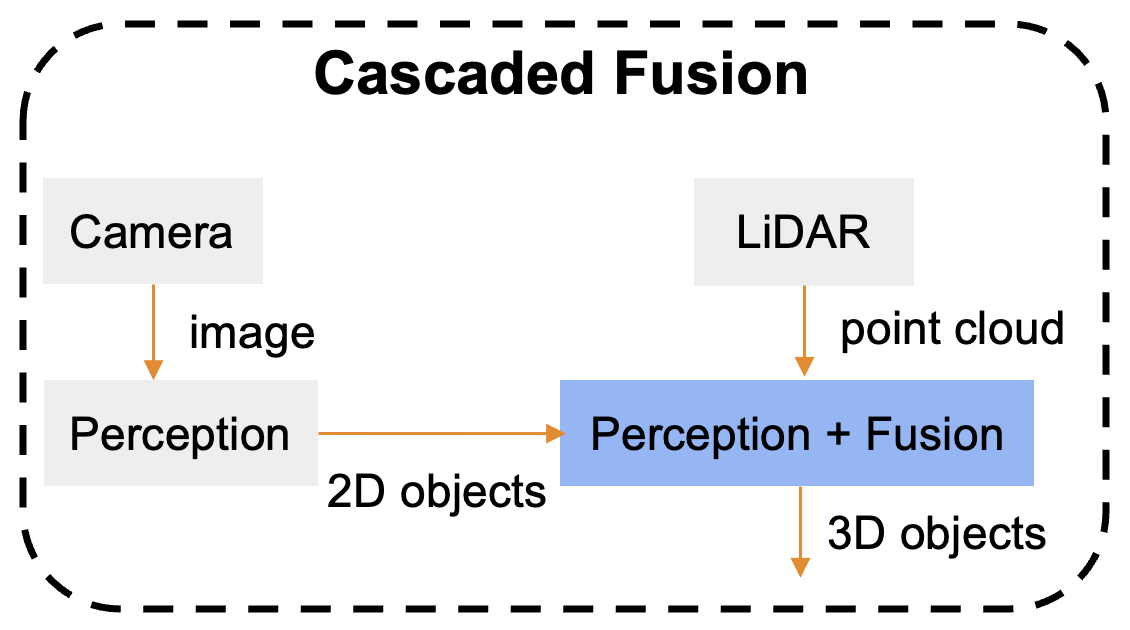}
    \caption{Cascaded fusion in autonomous vehicke systems.}
    \label{fig:cascaded_fusion}
\end{figure}

\noindent\textbf{Multi-Sensor Fusion Models}: To enhance the robustness of models and improve perception capabilities, sensor-fusion 3D object detection methods have been developed, leveraging multiple types of sensors with complementary characteristics. By integrating the unique strengths of different sensors, such as the depth information from LiDAR point clouds and the texture details from camera images, sensor fusion is recognized for achieving higher accuracy and greater robustness in detection tasks compared to using a single sensor alone. Given that autonomous vehicles are safety-critical systems, the reliability of object detectors is paramount. As a result, AVs often employ fusion detection models that combine data from both LiDAR and cameras, capitalizing on the complementary nature of these sensors. Based on the stage at which fusion occurs, multi-sensor fusion can be categorized into two main approaches: Cascaded Fusion and Parallel Fusion.

In \textit{\textbf{Cascaded Fusion}}, the integration of sensor data occurs in a sequential manner (Figure \ref{fig:cascaded_fusion}). Typically, one sensor's data, such as the camera's visual information, is processed first to identify key features or regions of interest. This processed data is then used to guide or refine the interpretation of the second sensor's data, like LiDAR, which provides depth and spatial information. The cascaded approach leverages the strengths of each sensor in stages, where the initial sensor can help focus or improve the accuracy of subsequent data interpretation. This method is beneficial for refining detections and reducing the computational load by narrowing down the areas for LiDAR analysis based on camera data.

For instance, PointFusion \cite{xu2018pointfusion} first processes the camera images to extract 2D region proposals. These proposals are then used to guide the processing of LiDAR point clouds. The 3D points are extracted only within the regions of interest (ROIs) identified by the camera, which reduces computational complexity. It allows the system to focus on specific areas identified as potential objects by the camera, making the subsequent LiDAR processing more efficient. MV3D (Multi-View 3D) \cite{chen2017multi} employs a cascaded fusion technique where it first uses 2D region proposals from camera images and then combines these proposals with LiDAR point clouds to generate 3D bounding boxes. The fusion occurs by projecting the 3D proposals onto bird's-eye view and front view for refinement. This method capitalizes on the strengths of both sensors by using camera data to guide the extraction of regions of interest from LiDAR data, improving 3D localization accuracy. 

\textit{\textbf{Parallel Fusion}} involves processing and integrating data from both LiDAR and cameras simultaneously. In this approach, the data from both sensors are processed in parallel, with each contributing its unique information to a unified perception model. 
Parallel fusion allows for a more holistic approach to perception, as it can consider the full data set from both sensors simultaneously, enhancing the system’s ability to detect and classify objects accurately in various conditions.

\begin{figure}
    \centering
    \includegraphics[width=\linewidth]{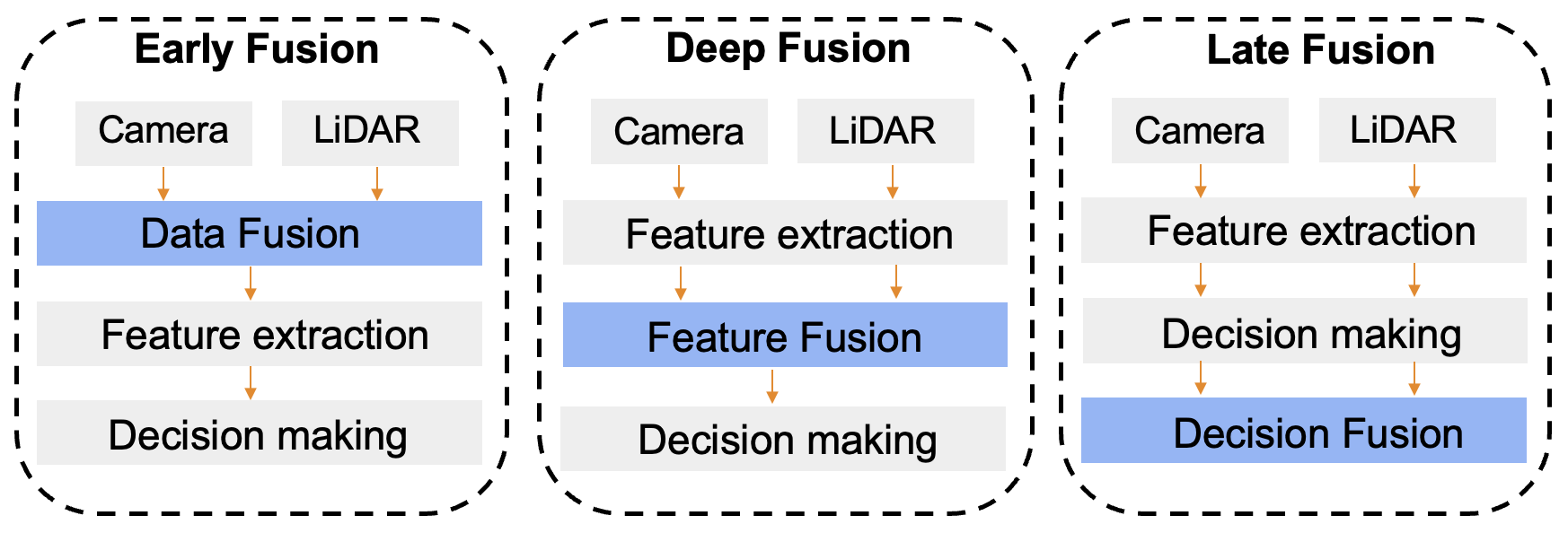}
    \caption{Various levels of parallel fusion in autonomous vehicle systems.}
    \label{fig:parallel_fusion}
\end{figure}

As shown in Figure \ref{fig:parallel_fusion}, parallel fusion can be performed at different stages: Input-Level Fusion (Early Fusion), Feature-Level Fusion (Deep Fusion), and Decision-Level Fusion (Late Fusion).

\noindent\textit{\textbf{- Input-Level Fusion (Early Fusion)}} is fusing data from multiple sensors right at the beginning of the processing pipeline before any significant feature extraction or processing takes place. This approach integrates different types of raw data early on, allowing the model to process and learn from this combined input throughout its layers.

\noindent\textit{\textbf{- Feature-Level Fusion (Deep Fusion)}} involves combining features extracted from different sensor modalities, such as point cloud data and image data, through various strategies (e.g., addition, averaging, or concatenation). These fused features are then fed into a detection network to obtain the final detection results. Examples of models that utilize this approach include AVOD \cite{ku2018joint}, and EPNet \cite{huang2020epnet}.

Qi et al. \cite{8578200} proposed F-PointNet, a cascading approach that first produces 2D proposals from camera images and projects them onto the LiDAR view as frustum proposals. The LiDAR point clouds within these frustum proposals are then used to generate corresponding 3D bounding boxes. Vora et al. \cite{vora2020pointpainting} introduced the PointPainting method, where LiDAR point clouds are projected into the output of an image-only semantic segmentation network. The segmentation scores are appended to each point, and the "painted" point clouds are subsequently fed into a LiDAR-only detector for 3D object detection. Huang et al. \cite{huang2020epnet} developed EPNet, which fuses the features of LiDAR point clouds and camera images on a point-wise basis. EPNet also introduces a consistency enforcing loss to address the inconsistency between classification confidence and localization accuracy, resulting in state-of-the-art performance for sensor-fusion-based 3D object detection. AVOD \cite{ku2018joint} employs a typical two-stage object detection architecture with a Region Proposal Network (RPN) followed by a second-stage detection network. It uses feature extractors to generate feature maps from both point clouds and images, which are then shared by two sub-networks. Initially, the feature maps are fed into the RPN and fused via an element-wise mean operation after cropping and resizing, generating the top k proposals through fully connected layers. These proposals are then projected onto the feature maps, where similar fusion operations as in the RPN stage are applied to produce the final detections, including box regression, orientation estimation, and category classification. In MVX-Net (Multi-View Fusion Network) \cite{sindagi2019mvx}, the point cloud data is first voxelized, and a 3D convolutional neural network (CNN) is used to extract features from the voxelized point cloud. Simultaneously, features are extracted from RGB images using a standard 2D CNN to identify key visual characteristics such as edges, textures, and colors. After independent feature extraction from the LiDAR and camera data, these features are fused at a deep level within the network. This fusion process typically involves projecting the 3D LiDAR features into the 2D image space, or vice versa, aligning the features based on spatial correspondences between the LiDAR points and image pixels, effectively combining the information from both modalities.

\noindent\textit{\textbf{- Decision-Level Fusion (Late Fusion)}} refers to the process of combining the results from separate detection networks, such as those based on LiDAR and camera data, after each modality has independently processed its input. The fusion occurs at the decision-making stage, where the outputs from both networks are merged using specific rules, such as geometric association or semantic consistency. This approach is exemplified by models like CLOCs \cite{pang2020clocs} and systems implemented in platforms like Apollo \cite{apollo} and Autoware \cite{autoware}. CLOC \cite{pang2020clocs} (Camera-LiDAR Object Candidates) is a model specifically designed for 3D object detection in autonomous vehicles, utilizing decision-level fusion to integrate object detection results from both camera and LiDAR modalities. In this approach, object detection is first performed independently on camera and LiDAR data. Typically, a 2D object detector (such as Faster R-CNN) processes the camera images to generate 2D bounding boxes, while a separate 3D object detector processes the LiDAR point cloud to generate 3D bounding boxes. The results from these independent detectors are then fused at the decision level, combining the strengths of both modalities to enhance the overall detection accuracy and robustness.

\subsubsection{Attack Form}

\begin{figure*}
    \centering
    \includegraphics[width=1\linewidth]{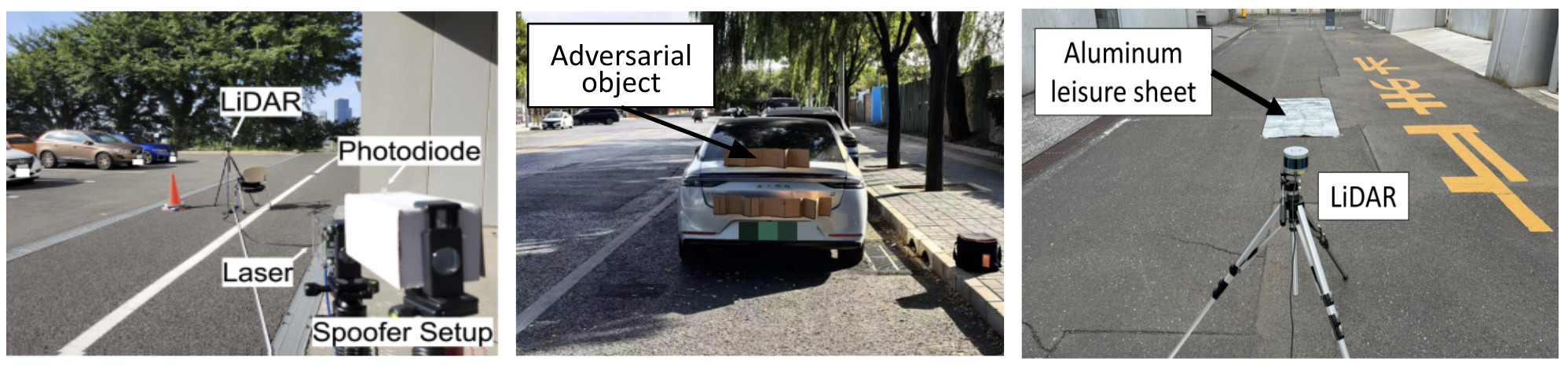}
    \caption{Illustration of different forms of physical attacks on LiDAR-based perception: \textit{Left:} Spoofing attack \cite{Sato2023}, \textit{Middle:} Adversarial object-based attack \cite{zhu2024}, and \textit{Right:} Reflective object-based attack \cite{Kobayashi2024}.}
    \label{fig:attack_forms}
\end{figure*}
\noindent\textbf{Spoofing Attacks}: Sensor spoofing attacks use the same physical channels as the targeted sensor to manipulate the sensor readings.  The adversary can achieve this by deploying a device within the line of sight of a victim vehicle’s LiDAR sensor. The adversarial device can capture LiDAR signals, alter them and emit them toward the victim sensor with a controlled delay. By controlling the return signal, the attacker can manipulate the resulting 3D measurements reported in a 3D point cloud by the victim sensor. This strategy makes it very difficult for the sensor system to recognize such attack, since the attack doesn’t require any physical contact or tampering with the sensor, and it doesn’t interfere with the processing and transmission of the digital sensor measurement. These types of attack could trick the victim sensor to provide seemingly legitimate but actually erroneous data.

\noindent\textbf{Physical Adversarial Objects:} These attacks involve designing physical objects with carefully crafted shapes that, when placed within the LiDAR sensor’s field of view, disrupt signal reflection. This manipulation results in misleading point clouds, which can cause the system to either misclassify or completely fail to detect objects. The adversarial objects are specifically designed to exploit weaknesses in the perception algorithms, making them appear as normal, yet altering the sensor's interpretation of the environment in critical ways.

\noindent\textbf{Reflective Objects}: Reflective objects can introduce errors in LiDAR perception by creating false or distorted point cloud data. Highly reflective surfaces, such as mirrors or polished metal, can reflect LiDAR signals in unexpected ways, leading to incorrect distance measurements or even ghost objects appearing in the LiDAR’s point cloud. These reflective materials can be strategically placed in an adversarial manner to disrupt the sensor’s ability to accurately perceive its surroundings.
\subsubsection{Placement in Physical Space}

\textbf{In Physical Proximity}: Locality in physical space refers to the necessity of an adversary being in close physical proximity to the target in order to carry out certain types of attacks. For many physical attacks, such as those involving direct manipulation of the environment or objects within the LiDAR sensor's field of view, the attacker must be near the target vehicle or sensor system. This proximity allows the attacker to place or alter objects, introduce reflective materials, or otherwise influence the physical environment in ways that the LiDAR system will misinterpret.

\noindent\textbf{On the Target Objects}: In addition to being in proximity, the attacker may need to focus specifically on the target objects themselves. This could involve adding a physical adversarial object on top of a target object, subtly altering its appearance or characteristics.

\section{Evaluations metrics on adversarial attacks
and model robustness }
\label{metrics}
We summarize evaluations metrics on adversarial attacks and model robustness in Figure \ref{fig:taxonomy}. 

\subsection{Attack Strength}





Evaluating the strength of an adversarial attack on LiDAR-based perception systems requires analyzing several performance metrics of the baseline 3D object detection models. These metrics help assess the effectiveness of the attack in degrading the performance of the target model.

\noindent\textbf{Recall-IOU Curve:}
The Recall-Intersection over Union (Recall-IOU) curve measures the detector’s recall for varying IOU thresholds. In the context of an adversarial attack, the goal is to reduce the model’s ability to correctly detect real objects. Recall is a critical metric in this scenario, as it reflects the model's ability to identify all relevant objects in a scene. A successful attack would aim to lower recall scores, indicating that the detector is missing objects it would otherwise recognize.

\noindent\textbf{3D Average Precision (3D AP):}
3D Average Precision (AP) is a key metric that evaluates the ratio of true positive predictions to all positive predictions. It assesses how well the detector performs across a range of confidence thresholds. In the context of adversarial attacks, reducing the 3D AP means that the attack has successfully caused the detector to misclassify or fail to detect objects. This is a primary measure of overall performance degradation in 3D object detection models.

\noindent\textbf{Average Confidence Score (ACS):}
The Average Confidence Score (ACS) reflects the detector's confidence in its predictions. An effective attack would reduce the ACS, causing the model to become less certain about its detections, even when objects are correctly identified. A significant drop in ACS suggests that the model's robustness has been compromised, leading to less reliable object detection.

\noindent\textbf{Detection Recall:}
Detection recall measures the percentage of true positives detected by the model out of all possible positive instances. A strong attack would aim to lower the detection recall by causing the model to miss objects that are present in the scene. A decrease in recall indicates the model’s reduced capacity to identify relevant objects, which can be particularly dangerous in safety-critical applications like autonomous driving.




\subsection{Attack Effectiveness}
\noindent\textbf{Attack Success Rate (ASR):}
The Attack Success Rate (ASR) measures the proportion of attacks that successfully deceive the target object detection model. Detection models typically apply default confidence score thresholds to filter out low-confidence detections, which may represent false positives. The ASR evaluates how effectively an attack fools the model by causing it to detect an object (e.g., a vehicle) with a confidence score that exceeds this threshold.


For example, in \cite{wang2023}, the ASR is calculated by determining the ratio of obstacles that were successfully detected with sufficient confidence to the total number of spoofed obstacles:

\begin{equation} 
ASR = \frac{\#~of~successful~detected~obstacles}{\#~of~total~spoofed~obstacles} 
\end{equation}

A high ASR indicates the attack's effectiveness in deceiving the model, while a lower ASR suggests the model is more resilient to adversarial inputs.

\noindent\textbf{Collision Rate:}
The collision rate is used as a critical evaluation metric, especially in the context of attacks aimed at object removal or misclassification in autonomous driving. This metric assesses how frequently the victim vehicle collides with an obstacle (e.g., another vehicle) due to the failure of the perception system to detect or correctly identify the object.

In \cite{Sato2023}, the collision rate is measured over multiple trials. For example, the collision rate is calculated as the number of times the victim vehicle collides with a sedan out of 10 trials:

\begin{equation} 
Collision~Rate = \frac{\#~of~collisions}{\#~of~trials} 
\end{equation}

\subsection{Attack Capability}


\textbf{a) For Spoofing Attacks}:\\
\noindent\textit{\textbf{Number of Spoofed Points:}} This indicates how many false points the attacker can inject into the LiDAR's point cloud data. A higher number of spoofed points allows the attacker to create more complex or convincing false objects in the perception system.

\noindent\textit{\textbf{Distance Control:}} The attacker’s ability to control the perceived distance of the spoofed points. By precisely manipulating the return time of the laser pulses, the attacker can make objects appear closer or further away than they actually are, potentially causing the vehicle to react inappropriately.

\noindent\textit{\textbf{Pattern Control:}} The ability to arrange the spoofed points into specific patterns. Pattern control can make the spoofed points form recognizable shapes or outlines, increasing the likelihood that the perception system will interpret them as specific objects, such as vehicles or pedestrians.

\noindent\textit{\textbf{Shape Control:}} The capability to create or modify the three-dimensional shape of the spoofed object. This involves arranging the spoofed points in a way that creates a convincing 3D shape, tricking the perception system into identifying it as a real object.

\noindent\textit{\textbf{Attack on Moving Vehicles:}} The ability to successfully carry out spoofing attacks on a moving target. This requires synchronizing the spoofed signal with the motion of the vehicle and maintaining the illusion in real-time, which is more complex than targeting a stationary vehicle.\\

\noindent\textbf{b) For Physical Adversarial Objects-based Attacks}:\\
\noindent\textit{\textbf{Shape Design:}} The ability to design physical objects with specific shapes that are crafted to distort the LiDAR point cloud data. These objects can be designed to manipulate the reflection and scattering of the laser beams, creating misleading 3D shapes that can confuse the perception algorithms into either misclassifying an object or failing to detect it altogether.

\noindent\textit{\textbf{Placement and Positioning:}} The capability to locate and strategically place the adversarial objects in the environment to maximize their disruptive effect. Proper positioning of these objects within the LiDAR's field of view can ensure they interact with the laser beams in a way that maximizes the chances of creating false positives, false negatives, or misleading data.

\noindent\textit{\textbf{Multi-View Consistency:}} The ability to create objects that can fool perception systems from multiple viewpoints. Since autonomous vehicles use LiDAR data from various angles to construct a 3D understanding of their surroundings, an effective adversarial physical object must maintain its deceptive properties across different viewpoints and sensor positions.

\noindent\textit{\textbf{Environmental Adaptability:}} The capability to adapt the attack to different environmental conditions such as lighting, weather, and occlusion. For instance, ensuring the adversarial object remains effective under various lighting conditions, or is designed to be effective even when partially occluded by other objects in the environment.\\


\textbf{c) For Reflective Objects-based Attacks}:\\
\noindent\textit{\textbf{Material and Surface Properties:}} The selection and use of materials with specific reflective or absorptive properties. By choosing materials that affect how the LiDAR beams are reflected back to the sensor, attackers can influence the intensity and distribution of the returned signals. This can cause the perception system to interpret the physical object differently or fail to detect it.

\noindent\textit{\textbf{Placement and Positioning:}} The capability to strategically place the adversarial objects in the environment to maximize their disruptive effect. Proper positioning of these objects within the LiDAR's field of view can ensure they interact with the laser beams in a way that maximizes the chances of creating false positives, false negatives, or misleading data.

\noindent\textit{\textbf{Environmental Adaptability:}} 
Reflective attacks can be adjusted to work in various lighting conditions, including day and night. Certain reflective materials may be more effective in specific environments, allowing attackers to optimize the attack based on the scenario.
\subsection{Attack Transferability}

Transferability refers to the phenomenon where adversarial attacks or perturbations that are effective against one model also demonstrate effectiveness against other models, even if those models differ in architecture, training methods, or datasets. This characteristic of adversarial attacks poses a significant security risk, as it suggests that a perturbation designed for one specific model may generalize and successfully deceive a wide range of models.

\noindent\textbf{Cross-model Transferability:}
In cross-model transferability, the adversarial perturbation is generated using one model (the training model) and is applied to a different model (the target model), which may have a different algorithm or architecture. Despite these differences, the attack remains effective, demonstrating that adversarial examples can generalize across models. This is particularly dangerous in real-world applications like autonomous driving, where attackers may not have access to the exact model but can still launch successful attacks on a variety of similar models.

\noindent\textbf{Cross-scene Transferability:}
Cross-scene transferability refers to the ability of an adversarial perturbation computed on one input point cloud to be successfully applied to a different point cloud. This implies that an adversarial perturbation crafted for a specific 3D scene can deceive the perception system in a new, unseen scene. The success of cross-scene attacks demonstrates that adversarial perturbations can generalize across different environments, making them even more dangerous in dynamic settings like autonomous driving, where the scene is constantly changing.

\subsection{Datasets} 
\textbf{KITTI dataset} is a popular dataset for benchmarking AD research, of which the point cloud data are by design divided into a trainval set containing 7481 samples and a test set containing 7518 samples. We follow the methodology by Chen et al. to split the trainval set to a training set (3712 samples) and a validation set (3769 samples) for better experimental studies \cite{chen20153d}. KITTI evaluates 3D object detection performance by average precision (AP) using the PASCAL \cite{everingham2010pascal} criteria and requires a 3D bounding box overlap (IoU) over 70\% for car detection. KITTI also deﬁnes objects into three difﬁculty classes: Easy, Moderate, and Hard. The difﬁculties correspond to different occlusion and truncation levels.

\noindent\textbf{Lyft dataset} known as the Lyft Level 5 Perception Dataset, is a large-scale dataset specifically designed for training and evaluating perception systems in autonomous vehicles. This dataset is part of the Lyft Level 5 Autonomous Vehicle Research initiative and is widely used in research focused on 3D object detection, sensor fusion, and autonomous driving in general.

\noindent \textbf{nuScenes dataset} is a comprehensive dataset specifically designed for autonomous vehicle research. Created by the autonomous driving company nuTonomy, which is part of Aptiv, the dataset has become one of the most popular and widely used resources in the AV research community, particularly for tasks related to 3D object detection, tracking, and sensor fusion.

\begin{table*}
    \centering
    \caption{Summary of Key Features and Attributes of different Datasets for Autonomous Driving Research}
    \begin{tabular}{|l|l|l|l|}
    \hline
    \textbf{Attribute/Feature} & \textbf{KITTI Dataset} & \textbf{Lyft Level 5 Perception Dataset} & \textbf{nuScenes Dataset}\\
    \hline
     Purpose    & Benchmarking for AD research &Training and evaluating perception & Autonomous vehicle research (3D object\\
         &  & systems in autonomous vehicles & detection, tracking, sensor fusion)\\
    \hline
     Origin    & Autonomous Driving Benchmark (KITTI) & Lyft Level 5 Autonomous Vehicle  & nuTonomy (part of Aptiv)\\
      &  & Research &  \\
    \hline
    Primary Focus     & 3D object detection & 3D object detection, sensor fusion, & 3D object detection, tracking, sensor \\
         &  & autonomous driving & fusion\\
    \hline
    Data Types  & Point clouds, images, annotations & Point clouds, images, annotations & Point clouds, images, annotations\\
    \hline
     Train/Validation/    & 7481 trainval (split into 3712 train, 3769  & Comprehensive training and  & Train/validation/test splits for \\
     Test Split    & val) and 7518 test samples & validation split; large-scale dataset & benchmarking\\
    \hline
     Evaluation Metric   & Average Precision (AP) using PASCAL  &Custom evaluation metrics for   & Average Precision (AP), 3D bounding \\
         & criteria, IoU > 70\% for car detection & perception system performance &  box IoU, multiple tasks\\
    \hline
     Difficulty Levels & Easy, Moderate, Hard & No explicit difficulty levels & No explicit difficulty levels \\
    \hline
     Unique    & - Widely used for 3D object detection & - Large-scale dataset & - Comprehensive dataset\\
      Characteristics   & - Defines three difficulty classes based on  & - Designed for AV perception tasks & - Widely used in AV research for 3D \\
         & occlusion and truncation &  & object detection, tracking, sensor fusion \\
    \hline
     Sensor Modalities    & LiDAR, cameras & LiDAR, cameras, RADAR & LiDAR, cameras, RADAR, GPS\\
    \hline
     Size    & 14,999 samples (7481 trainval, 7518 test) & Large-scale dataset (thousands of  & 1,000 scenes, each 20 seconds long; \\
         &  & samples) & 1.4 million camera images;\\
         &  &  & 390k LiDAR sweeps\\
    \hline
      Annotations   & 3D bounding boxes for objects  & 3D bounding boxes for objects & 3D bounding boxes for objects, object\\
         &  &  &  tracking information\\
    \hline
     Popular Tasks    &3D object detection, sensor fusion  & 3D object detection, sensor fusion & 3D object detection, tracking, sensor fusion\\
    \hline
    \end{tabular}
    \label{tab:datasets}
\end{table*}

Table \ref{tab:datasets} summarizes the features and attributes of the KITTI, Lyft Level 5, and nuScenes datasets. The choice of dataset should be aligned with the specific focus of the project:
KITTI Dataset is ideal for benchmarking tasks related to 3D object detection, particularly if the focus is on evaluating performance across varying levels of difficulty (Easy, Moderate, Hard). With its well-established use in the research community, it is a solid choice for comparative studies in 3D object detection and sensor fusion. Lyft Level 5 Perception Dataset is recommended for projects that require large-scale training and evaluation of perception systems. This dataset is particularly suited for research focusing on 3D object detection and sensor fusion in diverse urban environments, making it an excellent choice for developing and testing perception algorithms for autonomous vehicles. nuScenes Dataset offers comprehensive data across multiple modalities (LiDAR, cameras, RADAR, and GPS), making it highly valuable for research in 3D object detection, object tracking, and sensor fusion. Its large scale and detailed annotations make it ideal for researchers working on complex tasks involving sensor fusion and the evaluation of AV systems under various environmental and traffic conditions. For projects aiming to develop advanced perception and control systems, nuScenes provides a rich and diverse dataset.
\subsection{AD Simulators}

\textbf{LGSVL simulator} \cite{rong2020lgsvl} is a production-grade Autonomous Driving simulator based on the Unity 3D engine. It can perform environmental, sensor, and vehicle dynamics and control simulation of a vehicle. This capability allows users to customize environments and vehicles for testing and validation. The LGSVL simulator can interface with the Baidu Apollo platform \cite{apollo}, which is an open-source AV system that has over 100 partners and has reached multiple mass production agreements. The simulated vehicle in LGSVL can be controlled by Apollo in the virtual environment with perception, prediction, routing, and control modules. The newest Apollo 6.0 version updates its LiDAR-based perception module based on the PointPillar technique \cite{lang2019pointpillars}.

\noindent\textbf{CARLA simulator} \cite{dosovitskiy2017carla} is an open-source simulator for autonomous driving research. Developed to support the development, training, and validation of autonomous urban driving systems, CARLA provides a platform with high-fidelity simulation of urban environments, vehicles, and sensors. It supports a wide range of sensors, including cameras, LiDAR, radar, and GPS, making it suitable for a variety of autonomous driving research scenarios.
CARLA can be integrated with various autonomous driving stacks and machine learning frameworks. It supports ROS (Robot Operating System) for real-time communication and can interface with middleware platforms such as Autoware and Apollo.

\noindent\textbf{AWSIM simulator} \cite{AWSIM} is an advanced autonomous driving simulator designed to provide a high-fidelity environment for testing and developing autonomous vehicle technologies. It offers a realistic simulation platform that integrates various aspects of autonomous driving, including sensor data generation, environmental modeling, and traffic scenario management. By simulating complex driving environments with detailed physics and sensor models, AWSIM allows researchers and developers to evaluate the performance of autonomous systems in a controlled, repeatable, and safe manner. The simulator supports various sensor types, including LiDAR, cameras, and radar, enabling the testing of multi-sensor fusion algorithms. AWSIM also facilitates the development and testing of perception, planning, and control algorithms under diverse conditions, such as varying weather, lighting, and traffic scenarios. Its advanced capabilities make it a valuable tool for assessing the robustness and safety of autonomous driving systems, particularly in the context of adversarial attacks and defense mechanisms.

\begin{table*}
    \centering
    \caption{Comparison of Key Features of different Simulators for Autonomous Driving Research and Development}
    \begin{tabular}{|l|l|l|l|}
    \hline
     \textbf{Feature}	    & \textbf{LGSVL Simulator} & \textbf{CARLA Simulator} & \textbf{AWSIM Simulator}\\
    \hline
    Platform    & Unity 3D Engine & Custom Unreal Engine & Custom Engine\\
    \hline
     Primary Use Case    & Autonomous Vehicle simulation,  & Autonomous urban driving  & High-fidelity testing and evaluation of \\
         & sensor integration, vehicle control & research and development & autonomous driving systems\\
    \hline
     Sensors Supported    & Cameras, LiDAR, RADAR, GPS & Cameras, LiDAR, RADAR, GPS & Cameras, LiDAR, RADAR, GPS\\
    \hline
    Vehicle Dynamics    &Yes& Yes & Yes\\
    Simulation    & &  &\\
    \hline
     Environmental    & Customizable environments & Urban environments with & Detailed environmental \\
      Simulation   &  & high fidelity & and traffic modeling\\
    \hline
     Autonomous Stack   & Baidu Apollo & ROS, Autoware, Apollo & Multiple AV stacks, with focus on \\
     Integration      &  &  & robustness and safety\\
    \hline
    Middleware Support    & Baidu Apollo & ROS, Autoware, Apollo & ROS, Middleware support\\
    \hline
     Strength in Perception & LiDAR-based perception (PointPillar & High-fidelity perception,  & Robust perception testing, multi-sensor \\
         &   technique), detailed sensor simulations & multi-sensor fusion & fusion under diverse conditions\\
    \hline
    Planning and Control    & Integration with Apollo’s modules  & Supports path planning  & Comprehensive planning and control \\
         & (routing, prediction, control) & and behavior planning & evaluation with environmental variations\\
    \hline
    Adversarial Attack & Limited & Moderate & Advanced, with capabilities  \\
   Testing    &  &  &for simulating complex scenarios  \\
                  &  &  & and defense mechanisms \\
    \hline
    Weather \& Lighting & Limited & Supports weather and & Supports various weather, lighting, \\
    Conditions      &  & lighting changes & and traffic scenarios\\
    \hline
    Customization   & High customization of environments & Moderate customization for  & High customization for testing\\
         & and vehicles & urban driving scenarios & conditions and sensor configurations\\
    \hline
   Community \& Ecosystem     & Large community & Extensive research community & Emerging, with focus on detailed\\
         & (Baidu Apollo ecosystem) & open-source & testing for research and development\\
    \hline
     Best Suited For    &Testing and validating AV systems & Research and development & High-fidelity testing of AV perception,\\
         & with Apollo integration, sensor fusion &  of urban driving  & planning, and control under diverse\\
         &   & systems, multi-sensor fusion, & and adversarial conditions \\
        &   &  integration with ML frameworks  &  \\
    \hline
    \end{tabular}
    \label{tab:simulators}
\end{table*}

Table \ref{tab:simulators} summarizes the key features of the LGSVL, CARLA, and AWSIM simulators, highlighting their strengths and the types of work they are best suited for:
LGSVL Simulator is best for projects involving the Baidu Apollo platform or those that require customizable environments with a focus on sensor integration and vehicle control. Ideal for testing perception, prediction, and routing modules using specific autonomous driving platforms like Apollo. CARLA Simulator is suitable for research and development of autonomous urban driving systems, particularly when high-fidelity simulation of urban environments, sensor diversity, and integration with machine learning frameworks is required. CARLA's support for various autonomous driving stacks makes it a versatile choice for academic and research purposes. AWSIM Simulator is recommended for projects requiring high-fidelity simulations in diverse environmental and traffic conditions. AWSIM is particularly valuable for evaluating the robustness and safety of autonomous driving systems, including testing against adversarial attacks and defense mechanisms.
\subsection{AD Platforms}

\textbf{Baidu Apollo platform} \cite{apollo} is an open-source AV system that has reached multiple mass production agreements. The Apollo technology stack has been in various levels of testing since 2017 and has attracted over 100 collaborators, including leading automakers like Toyota, Geely, Daimler, BMW, Hyundai, and Ford as well as other industry partners such as Nvidia, Bosch, Intel, and TomTom.

\noindent\textbf{Autoware platform} \cite{autoware} is an open-source software stack for self-driving vehicles, built on the Robot Operating System (ROS). It includes all of the necessary functions to drive an autonomous vehicles from localization and object detection to route planning and control, and was created with the aim of enabling as many individuals and organizations as possible to contribute to open innovations in autonomous driving technology.

\label{techniques}

\section{Design Challenges and Physical Constraints}

\subsection{Attacks on Images vs. Point Clouds}

Attacking images and point clouds presents unique challenges due to the differences in their data structures and the methodologies employed.

\begin{figure}
    \centering
    \includegraphics[width=1\linewidth]{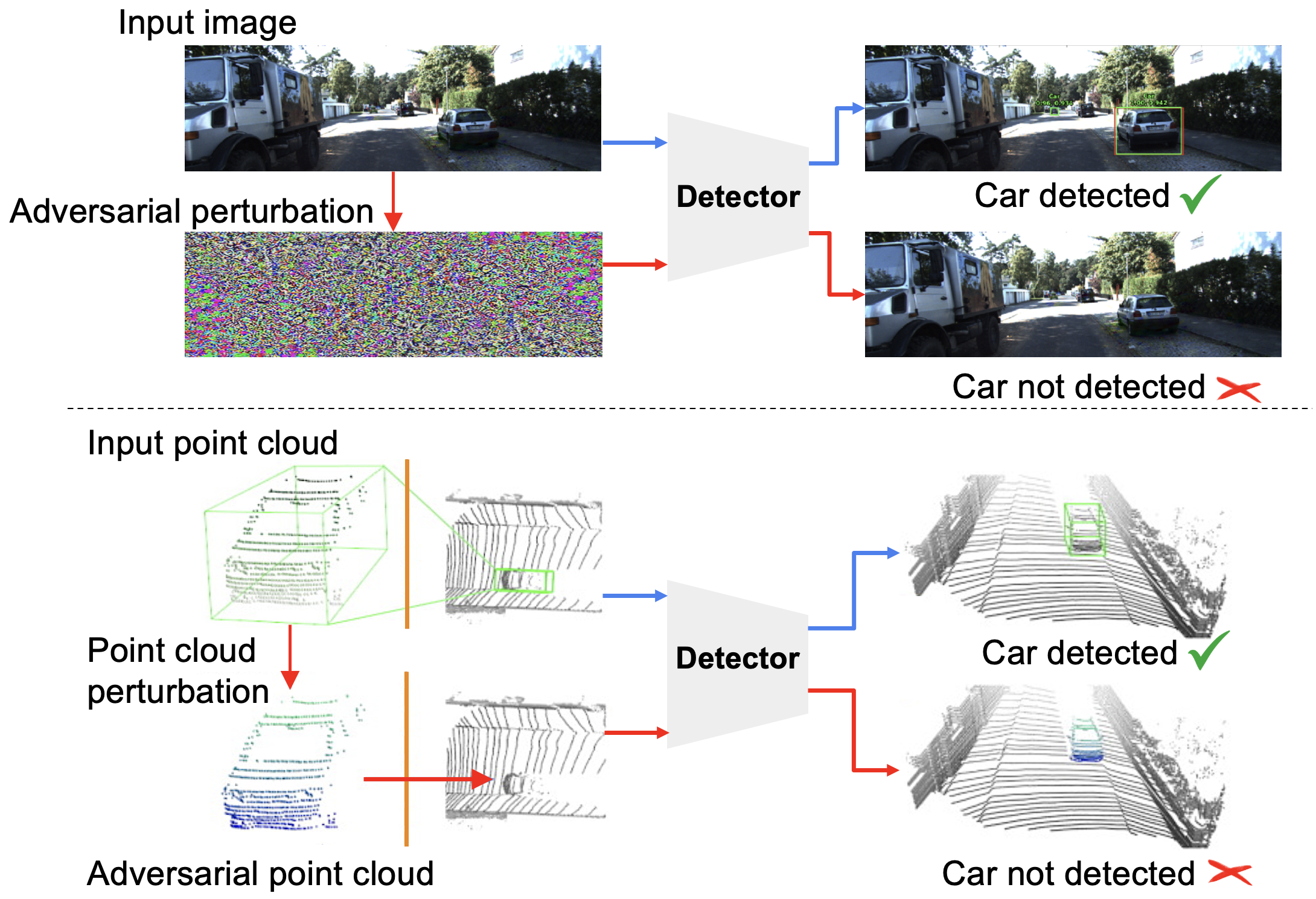}
    \caption{Illustration of attacks on image vs. point cloud.}
    \label{fig:imagevspoint}
\end{figure}

\noindent\textbf{Different Perturbation Methods:}
Images and point clouds differ fundamentally in how data is structured and manipulated (see Figure \ref{fig:imagevspoint}). Images have a compact, ordered structure, while point clouds are irregular, represented as $N \times C$, where $N$ denotes the number of points and $C$ includes spatial and intensity data (e.g., xyz-i). In image-based attacks, adversaries typically manipulate RGB values to create adversarial examples. On the other hand, in LiDAR-based attacks, adversaries directly manipulate the positions of points in 3D Euclidean space while ensuring the perturbations conform to the physical constraints of the LiDAR system.

\noindent\textbf{Different Perturbation Capabilities:}
In 2D image attacks, the entire target surface can often be used for the attack, assuming the adversary has full control (e.g., manipulating any part of a stop sign, as shown in \cite{eykholt2018robust}). However, in LiDAR spoofing attacks, the adversary's attack surface is significantly limited by the physical capabilities of the sensor. This smaller attack surface makes it more challenging to execute effective attacks on point clouds compared to images.

\noindent\textbf{Different Perturbation Constraints:}
Image-based attacks typically employ Lp norms to constrain the perturbations, with the objective of making them imperceptible to human observers \cite{eykholt2018robust}. In contrast, LiDAR attacks do not prioritize stealthiness in the same way, as point clouds are not visually interpreted by humans. The main constraint in LiDAR-based attacks is ensuring that perturbations remain within the physical and sensor-based limitations of the LiDAR system, which becomes the primary factor in determining the attack’s success.

\subsection{Design Challenges for Multi-Sensor Based Perception}

\noindent\textbf{Unified Physical-World Attack Vector for Both Camera and LiDAR:}
Finding a single attack vector that can deceive both camera- and LiDAR-based perception systems is a significant challenge. Existing methods often require separate attack vectors, such as stickers or lasers, which increase the complexity and cost of attacks. A unified attack vector would streamline the process and improve stealthiness.

\noindent\textbf{Differentiable Synthesis of Attack Impacts on Both Modalities:}
Generating adversarial inputs requires a differentiable approach to synthesize physical-world impacts on both camera and LiDAR data. This is critical for optimization-based attack generation, where repeated iterations are necessary. A differentiable model allows for more efficient and effective optimization without needing to physically test the attack after each iteration.

\noindent\textbf{Handling Non-Differentiable Pre-Processing Steps in AD Systems:}
In practical AD systems, both camera images and LiDAR point clouds are pre-processed before being fed into multi-sensor fusion algorithms. However, some pre-processing steps, like aggregating 3D points.

\subsection{Spoofing Attacks}

\subsubsection{Object Injection Constraints}

LiDAR-generated point clouds are inherently sparse, as each laser ray can capture only one point. To ensure adversarial points emitted by an attacker's transmitter are detected by the LiDAR, they must align with the sensor's laser rays. Several physical constraints must be observed when generating these adversarial points:

\begin{itemize} 
    \item \textbf{Single point per laser ray:} In many autonomous vehicles, the LiDAR operates in the Strongest Return Mode \cite{cao2019ccs}, where only the point with the strongest reflection is recorded. If adversarial points are not aligned correctly, multiple points may appear on the same laser ray, violating the physical constraints of the LiDAR.

    \item \textbf{Vertical alignment with discrete laser angles:} Mechanical LiDARs used in autonomous driving often have preset configurations (e.g., 16-line, 32-line, or 64-line systems). The adversarial points must be placed within these discrete vertical angles, corresponding to the LiDAR's beams.

    \item \textbf{Horizontal angle limitations:} The adversary’s capability to inject points is constrained by hardware limitations, such as the laser transmitter. Injecting points across the full 360° horizontal field of view is not feasible, and current attack methods typically work within a limited horizontal range of about 10°.
\end{itemize}

\subsubsection{Object Removal Constraints}

Object removal attacks face two key limitations:

\begin{itemize} 
    \item \textbf{Limited receptive field of the LiDAR’s photodiodes:} LiDAR systems are designed to receive reflections from specific directions. If the spoofer falls outside this receptive range, the LiDAR fails to detect the spoofed signals, rendering the removal attack ineffective.

    \item \textbf{Limited output power of the laser diodes in a single-spoofer setup:} The output power of the laser diodes determines the range and precision of the spoofed pulses. As the laser beam spreads, the intensity of the signal decreases, particularly near the edges, leading to a weaker signal that may be insufficient for the LiDAR to register. Furthermore, as the distance between the spoofer and LiDAR increases, the signal decays, reducing the attack's overall effectiveness.

\end{itemize}

\subsection{Physical Object-Based Attacks}

Converting adversarial point clouds into real-world physical objects introduces additional challenges, particularly when transferring these "virtual points" into actual road conditions.

\subsubsection{Technique 1: Targeting LiDAR (Shape Only)}

Tsai et al. \cite{tsai2020robust} address this by reconstructing surfaces from adversarial point cloud data. However, the uncertainty in the reconstruction process reduces the attack’s success rate in real-world settings. Their method also overlooks specific sensor mechanisms and relies on random sampling, which limits the accuracy of the simulation.

In \cite{yang2021}, physical objects are designed to appear as adversarial point clouds when scanned by LiDAR. Their method first simulates the LiDAR’s scanning process using 3D meshes, creating a point cloud by calculating intersections between LiDAR rays and object surfaces. Key parameters, such as the LiDAR’s position and resolution, can be adjusted to match the real-world configuration.

The simulation process involves: \begin{itemize} \item \textbf{Line-plane intersection:} Calculating the intersections between LiDAR rays and object surfaces. \item \textbf{Point-in-polygon check:} Verifying whether the intersection points lie within the object’s polygons (surfaces). \item \textbf{Distance comparison:} Filtering out points obstructed by other objects. \end{itemize}

\noindent\textbf{Physical Printing Constraints}
When deploying adversarial objects in real-world environments, physical constraints such as size and stability must be considered. 3D printers limit the size of adversarial objects to 45 cm × 45 cm × 41 cm, which is much smaller than typical vehicles or pedestrians. A flat surface could be added to the object’s base to ensure stability. In \cite{yang2021}, the goal is to mislead the model, so the resemblance between the original and adversarial objects is not a priority. This allows flexibility in shaping the adversarial object without strict distance constraints, such as the $L_2$ norm or Chamfer distance.

\subsubsection{Technique 2: Targeting LiDAR and Camera (Shape and Texture)}

A more complex attack targets both LiDAR and camera-based perception systems. A mesh representation is used to maintain realistic 3D geometry and generate adversarial textures that affect both RGB images and point clouds. The mesh is positioned on top of a vehicle and aligned with its orientation.

The adversarial object’s shape is trained by modifying the vertices of an initial mesh using learnable displacement vectors, which are applied to each vertex. A transformation matrix is used to position and orient the mesh on the vehicle.

To ensure realism, constraints on size and smoothness are imposed on the mesh geometry, and smooth texture transitions are generated by interpolating between vertex colors.

\noindent\textbf{Differentiable Rendering:}
For accurate rendering, the LiDAR’s rays are simulated, and the intersection points with the mesh’s triangles are calculated using the Möller–Trumbore intersection algorithm \cite{moller2005fast}. The nearest intersection points are added to the point cloud. The mesh is also rendered into 2D to optimize the adversarial texture using differentiable rendering techniques \cite{johnson2020accelerating}.

\noindent\textbf{Physical Printing Constraints:}
Adversarial objects are created with an isotropic sphere mesh and box constraints to maintain their original dimensions. The ADAM optimizer is used to iteratively deform the mesh and minimize the attack objective function. The final adversarial shape is then rendered into 2D images, and a universal adversarial texture is trained for effective attacks on both RGB and point cloud inputs.

\section{Attacks on LiDAR-based Perception} 
\label{lidar_perception}

Few attacks targeting the perception modules of autonomous vehicle systems consider both LiDAR sensors and the deep learning models that process their data simultaneously. While adversarial attacks on 3D point cloud data have been proposed \cite{liu2019extending, xiang2019}, demonstrating the feasibility of altering point cloud data to deceive deep learning models, these approaches often fall short of practical real-world implementation.

In this section, we will discuss various attacks targeting LiDAR-based systems, as outlined in Table \ref{tab:attacks_main}. These attacks exploit the inherent vulnerabilities of LiDAR sensors and their underlying algorithms, demonstrating different approaches to compromise the reliability of LiDAR-based perception in autonomous systems. The table includes attacks that range from simple signal interference to complex adversarial examples designed to manipulate the LiDAR-generated point clouds. Each attack method highlights specific weaknesses in how LiDAR data is processed, whether through spoofing techniques or adversarial objects. 

\begin{table*}[!htp]
\centering
  \caption{Main adversarial attack methods against only LiDAR-based systems: Attack Setting, Attacker Knowledge, Attack Goal (OI-Object Injection; OR-Object Removal; T-Translation; MC-Miscategorization), Attack Scenario, Attack Form, Perception and Venue.}
  \label{tab:attacks_main}
  \begin{tabular}{|l|c|c|c|c|c|c|}
    \hline
       \textbf{Attack}  & \textbf{Setting} &\textbf{Knowledge} & \textbf{Goal} &  \textbf{Form} &\textbf{Perception} & \textbf{Venue}  \\
    \hline 
       Cao et al.\cite{cao2019ccs} & Physical  & White-box  & OI  & Spoofing Attack & Single-Modality & CCS 2019\\
       \hline
       Cao et al. \cite{cao2019adversarial} & Physical  & Black-box/White-box & OR, MC  & Physical Adversarial Object  & Single-Modality & Arxiv 2019\\ \hline      
       Sun et al. \cite{sun2020} & Physical  & Black-box & OI  & Spoofing Attack & Single-Modality & USENIX 2020\\
       \hline
       Tu et al. \cite{tu2020} & Physical  & Black-box/White-box & OR  & Physical Adversarial Object & Single-Modality& CVPR 2020\\
       \hline
      Yang et al. \cite{yang2021} & Physical  & White-box/Black-box & OI & Physical Adversarial Object & Single-Modality & ASIA-CCS 2021\\
      \hline
       Zhu et al. \cite{zhu2021} & Physical   & Black-box & OR  & Reflective Objects & Single-Modality & CCS 2021\\
       \hline
       Wang et al. \cite{wang2023} & Physical   & Black-box/White-box & OI &  Spoofing Attack & Single-Modality & IEE TMM 2023\\
       \hline
       Cao et al. \cite{cao2023} & Physical  & Grey-box & OR & Spoofing Attack & Single/Multi-Modality & USENIX 2023\\
       \hline
       Jin et al. \cite{jin2023} & Physical  & White-box & OI,OR & Spoofing Attack & Single-Modality & S\&P 2023\\
       \hline
        Sato et al. \cite{Sato2023} & Physical  & Black-box & OI, OR &  Spoofing Attack & Single-Modality & NDSS 2024\\
       \hline
        Suzuki et al. \cite{suzuki2024}   &  Physical & White-box & OR  &Spoofing Attack & Single-Modality & VehicleSec 2024 \\
       \hline
        Zhu et al. \cite{zhu2024}   & Physical  & White-box & OR & Physical Adversarial Object & Single-Modality & USENIX 2024 \\
       \hline
       Kobayashi et al. \cite{Kobayashi2024} & Physical  & White-box & OI  & Reflective Objects & Single-Modality & NDSS Symp. 2024\\
 \hline
\end{tabular} 
\end{table*}


\begin{figure}
    \centering
    \includegraphics[width=1\linewidth]{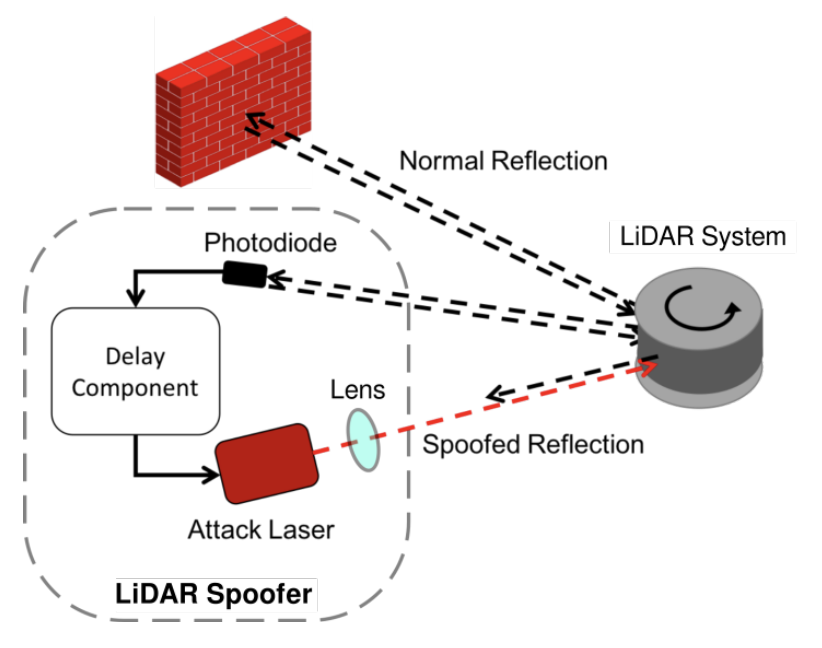}
    \caption{Illustration of a LiDAR spoofing attack. The photodiode captures laser pulses from the LiDAR, which activates a delay component that triggers the attacker's laser to emit simulated echo pulses, mimicking real object reflections (Figure adapted from \cite{cao2019ccs}).}
    \label{fig:cao_spoofing}
\end{figure}


Cao et al. \cite{cao2019ccs} were the first to explore the security vulnerabilities of LiDAR-based perception in AV systems. However, their laser-spoofing attack was only tested in a simulation environment. The attack requires precise dynamic aiming of the spoofing device at the LiDAR sensor on the target vehicle, see Figure \ref{fig:cao_spoofing}, making it highly impractical to execute in real-world driving scenarios. The authors adopted an optimization-based approach to find adversarial points by minimizing an adversarial loss function, tailored to the functionality of the machine learning model being attacked. The Adam optimizer was used to solve the corresponding optimization problem, in combination with global sampling to mitigate the issue of getting trapped in local optima—a common limitation of gradient-based methods like Adam. While this approach is effective, it has a major drawback: it requires knowledge of the underlying model, making it less applicable in black-box scenarios where the attacker's access to model details is limited.

In contrast, Sun et al. \cite{sun2020} approached this problem from a black-box perspective, which is more aligned with real-world scenarios where attackers typically do not have direct access to the target model. This approach avoids the need for detailed knowledge of the model, making it more feasible for practical attacks in real AV settings. Though Sun et al. extended the work into a black-box setting, the practical applicability of their attack in real road environments remains limited. Sun et al. observed that LiDAR sensors capture only a sparse set of points from partially occluded vehicles (e.g., cars hidden by other vehicles) or those located at greater distances. To exploit this, they incorporated point clouds from partially occluded and distant cars, using both real LiDAR measurements and artificially generated data with a 3D renderer. These point clouds were then injected into clean LiDAR data, resulting in a high success rate for the attack, with point cloud classifiers detecting fake vehicles in the immediate vicinity.

However, despite the attack's effectiveness in controlled conditions, its practicality in dynamic, real-world driving environments is still questionable. The complexity of accurately positioning fake point clouds in rapidly changing road scenarios and ensuring consistent, undetectable deception across diverse driving conditions remains a significant challenge.

\begin{figure}
    \centering
    \includegraphics[width=1\linewidth]{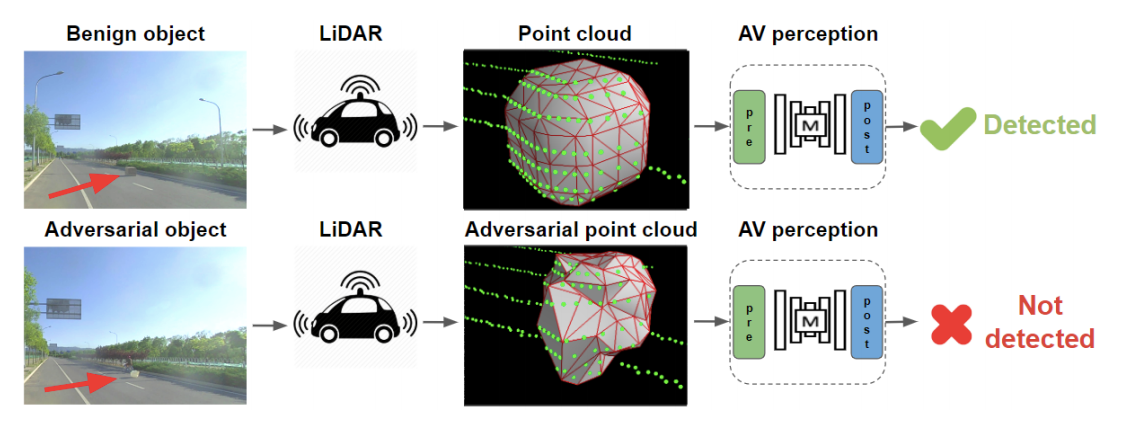}
    \caption{Overview of LiDAR-Adv. In the first row, a standard object is successfully detected by the LiDAR-based detection system. However, in the second row, the adversarial object—despite being of a similar size—evades detection (Figure adapted from \cite{cao2019adversarial}).}
    \label{fig:cao_adversarial}
\end{figure}

Towards generating physically realizable attacks, Cao et al. \cite{cao2019adversarial} proposed a technique named \textit{LiDAR-Adv} to learning an adversarial mesh capable of generating adversarial point clouds using a LiDAR renderer (see Figure \ref{fig:cao_adversarial}). However, their approach was limited to a few specific frames, meaning the learned 3D object was not universal and could not be effectively reused in other 3D scenes. Additionally, their evaluation was conducted on a small, in-house dataset containing only a few hundred frames, which limits the generalizability of their results. While they made initial attempts to execute the attack in real-world environments, the lack of detailed information—such as code, algorithms, and sufficient experimental results—left gaps in their study.

Creating effective adversarial samples for LiDAR-based detection systems presents several challenges:
\begin{enumerate}
    \item \textit{Uncertainty in Shape Perturbations:} LiDAR-based detection systems use solid-state LiDAR devices to convert a 3D shape into a point cloud, which is then processed by machine learning models. The relationship between shape perturbations and their effects on the scanned point cloud is not straightforward.
    \item \textit{Incompatibility with Gradient-Based Optimization:} Traditional gradient-based optimizers struggle with the pre-processing steps involved in LiDAR point cloud data, making them ineffective for this task. A new optimization method is needed that works directly with the point cloud.
    \item \textit{Restricted Perturbation Space:} The perturbation space in which adversarial attacks can operate is constrained by both physical limitations (e.g., real-world realizability) and the characteristics of the LiDAR sensor itself.
\end{enumerate}

To overcome these challenges, researchers first developed a differentiable LiDAR renderer that could link 3D target perturbations to the resulting point cloud. This enabled them to simulate how changes to the 3D object would affect the LiDAR scan in a differentiable manner. They then performed 3D feature aggregation using a proxy function that allowed for gradient-based optimization. Finally, they designed specific loss functions to ensure the generated adversarial 3D samples were smooth and realistic.

The results demonstrated that, by leveraging 3D sensing technologies and multi-stage object detectors commonly used in autonomous vehicles, researchers could successfully mislead the perception systems of autonomous driving systems. However, more robust real-world testing and broader datasets are needed to further validate these attacks.

\begin{figure}
    \centering
    \includegraphics[width=1\linewidth]{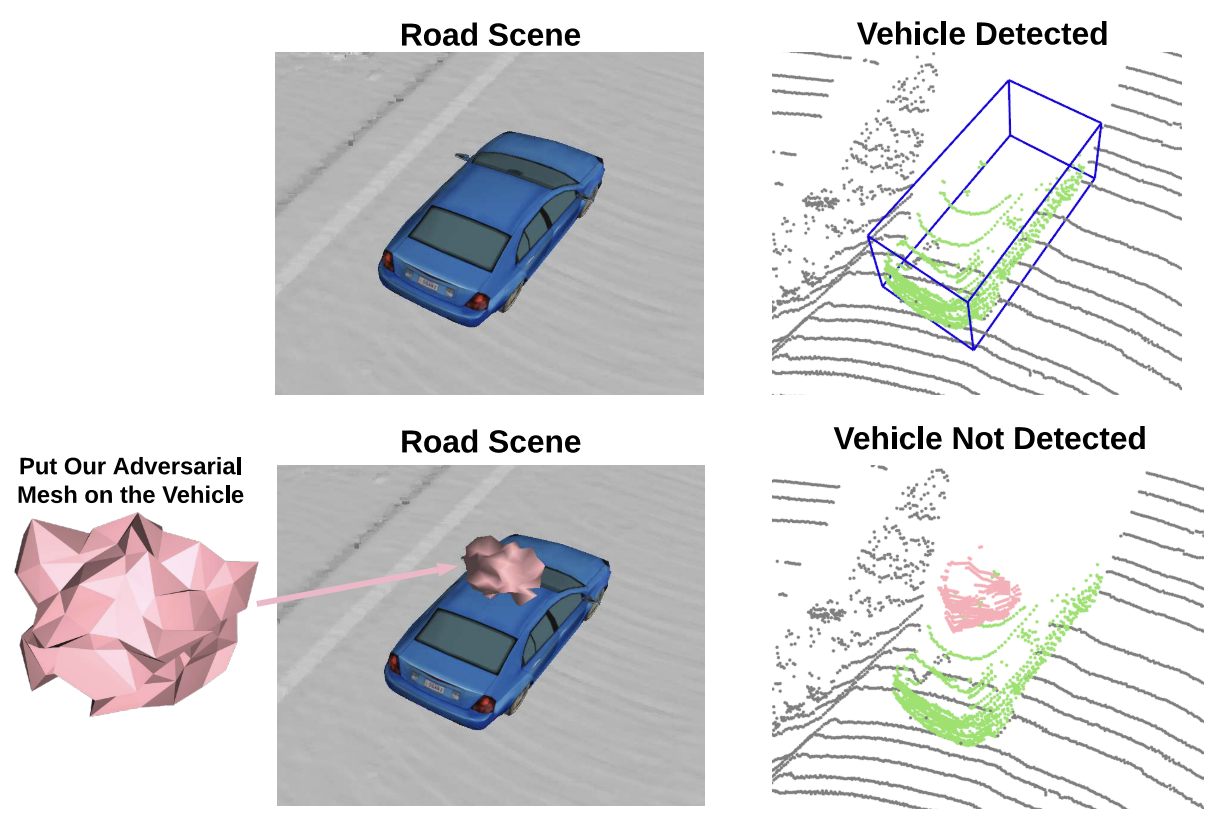}
    \caption{A physically realizable adversarial object designed to render vehicles \textit{invisible}. When placed on the rooftop of a target vehicle, it prevents the vehicle from being detected by the LiDAR-based detection system (Figure adapted from \cite{tu2020}).}
    \label{fig:tu2020}
\end{figure}

Another significant contribution comes from Tu et al. \cite{tu2020}, who present a physically realizable adversarial example, such as one that can be fabricated using a 3D printer. When this object is mounted on the roof of a car, as illustrated in Figure \ref{fig:tu2020}, it effectively hides the vehicle from LiDAR-based perception systems used by other autonomous vehicles. Tu et al. proposed both white-box and black-box methods to generate these adversarial objects, achieving an 80\% success rate in evading point-cloud-based object detectors when placed above a target vehicle.

For the white-box attack, a gradient-based approach is used to generate the adversarial object by minimizing the confidence score of the target vehicle, effectively rendering it undetectable. In the black-box scenario, Tu et al. demonstrated an approach where adversarial objects are generated using a genetic algorithm. This algorithm iteratively evolves the mesh of the object to improve its ability to evade detection, without requiring detailed knowledge of the object detection model.

The significance of these object-hiding attacks cannot be overstated. While spoofing attacks might lead to unnecessary stops, as the autonomous vehicle detects a non-existent object, object-hiding attacks are far more dangerous. Failing to detect an actual object, such as a vehicle, increases the likelihood of fatal collisions. This makes object-hiding attacks a critical threat to autonomous driving systems.

Yang et al. \cite{yang2021} propose an attack that generates adversarial 3D point clouds against deep learning models in both white-box and black-box scenarios. This includes the creation of robust physical adversarial objects placed at the roadside, which are detected by LiDAR sensors as vehicles, potentially causing traffic jams, emergency stops, or irregular lane changes. 

For the white-box attack, authors assume that  the attacker has access to the target deep learning model (PointRCNN) and generates adversarial point clouds by manipulating vertices in 3D object meshes. As for the black-box attack, they target models like PointPillar and PV-RCNN using a genetic-evolving algorithm to generate adversarial objects without access to model internals.

\begin{figure*}
    \centering
    \includegraphics[width=0.8\linewidth]{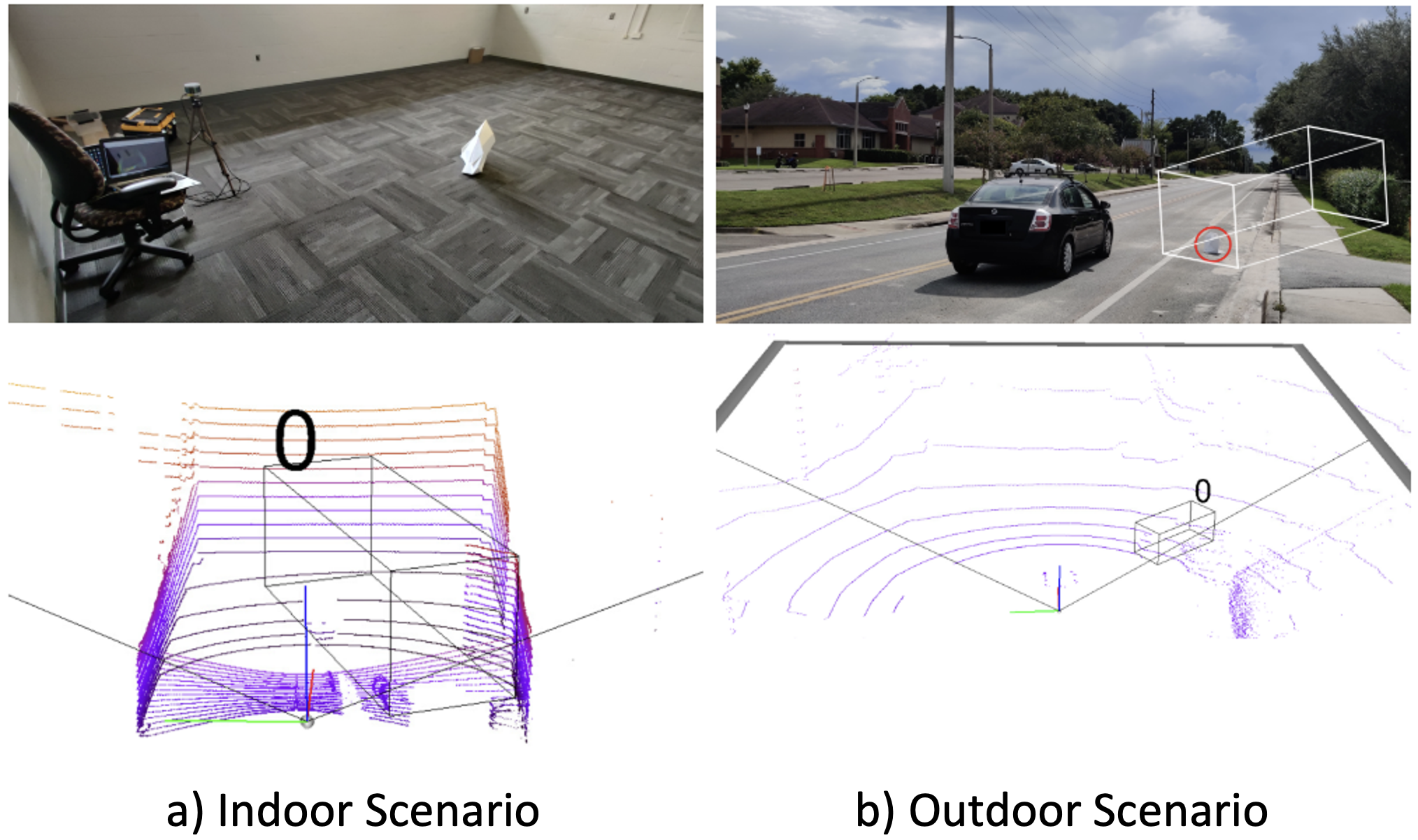}
    \caption{Visualization of appearing attack in indoor and outdoor scenarios (Figure adapted from \cite{yang2021}).}
    \label{fig:yang2021}
\end{figure*}

The attack is implemented with 3D printed adversarial objects placed roadside, demonstrating the potential impact on real-world driving scenarios (see Figure \ref{fig:yang2021}). The paper tests the effectiveness of the attack using both simulated environments (LGSVL simulator with Baidu Apollo) and real road tests.

The paper evaluates existing defense mechanisms against adversarial point clouds and shows that their proposed attack can bypass these defenses with high success rates. They also propose a new detection method to identify physical adversarial objects based on the physical characteristics of current LiDAR sensors. The attack successfully misleads deep learning models to perceive small roadside objects as vehicles. In simulation, it caused abnormal driving behaviors such as sudden stops or lane changes. The attack bypasses existing defenses like outlier point removal and random Gaussian noise addition.

Zhu et al. \cite{zhu2021} propose a novel attack framework called \textit{AdvLo} where the attacker identifies specific adversarial locations in physical space. By placing arbitrary reflective objects at these locations, they can fool the LiDAR perception systems, achieving a success rate of over 90\%. The attack's feasibility is demonstrated using commercial drones, making it the first study to use such an approach. The attack framework consists of two main steps: (i) Location Probing: This step involves finding a large number of locations with a high probability of being adversarial. The authors use a novel algorithm to probe these locations efficiently. (ii) Location Selection: Once potential adversarial locations are identified, the most critical ones are selected based on their impact on the LiDAR perception system's outputs. The selection is guided by an adversarial score that measures the negative effect of each location on the system's detection accuracy.

In their real-world experiment, the authors used two drones to hover at the critical adversarial locations identified by their framework. The drones' reflective surfaces created sufficient interference in the LiDAR system, causing it to fail in detecting a car in front of the autonomous vehicle. The study also demonstrated that the attack could be sustained as the victim vehicle approached the target, showing the method's potential for causing persistent misperception in real-world scenarios.

The authors also discuss possible defense mechanisms to counteract such attacks, such as enhancing the robustness of LiDAR perception systems against adversarial perturbations and developing strategies for detecting abnormal point cloud patterns that indicate an attack.

Wang et al. \cite{wang2023} introduce an adversarial attack algorithm that uses physical LiDAR simulation to construct sparse obstacle point clouds and then adversarially perturbs prototype points along each ray direction to deceive 3D detection models. Experiments showed that voxel-based detectors are more vulnerable to these adversarial attacks than point-based methods. The algorithm achieves an 89\% mean attack success rate against the PV-RCNN detector using only 20 points to spoof a fake car. Their methodology is based on two stages: 
\begin{itemize}
    \item Stage 1: The algorithm first generates sparse obstacle point clouds by simulating a LiDAR sensor to construct a target object's shape, integrating it naturally into the scene. 
    \item Stage 2: It then perturbs these points in the polar coordinate system to increase their deception ability against 3D detectors. This perturbation is constrained along the ray direction, adhering to the physical limitations of LiDAR sensors.
\end{itemize}

\begin{figure}
    \centering
    \includegraphics[width=1\linewidth]{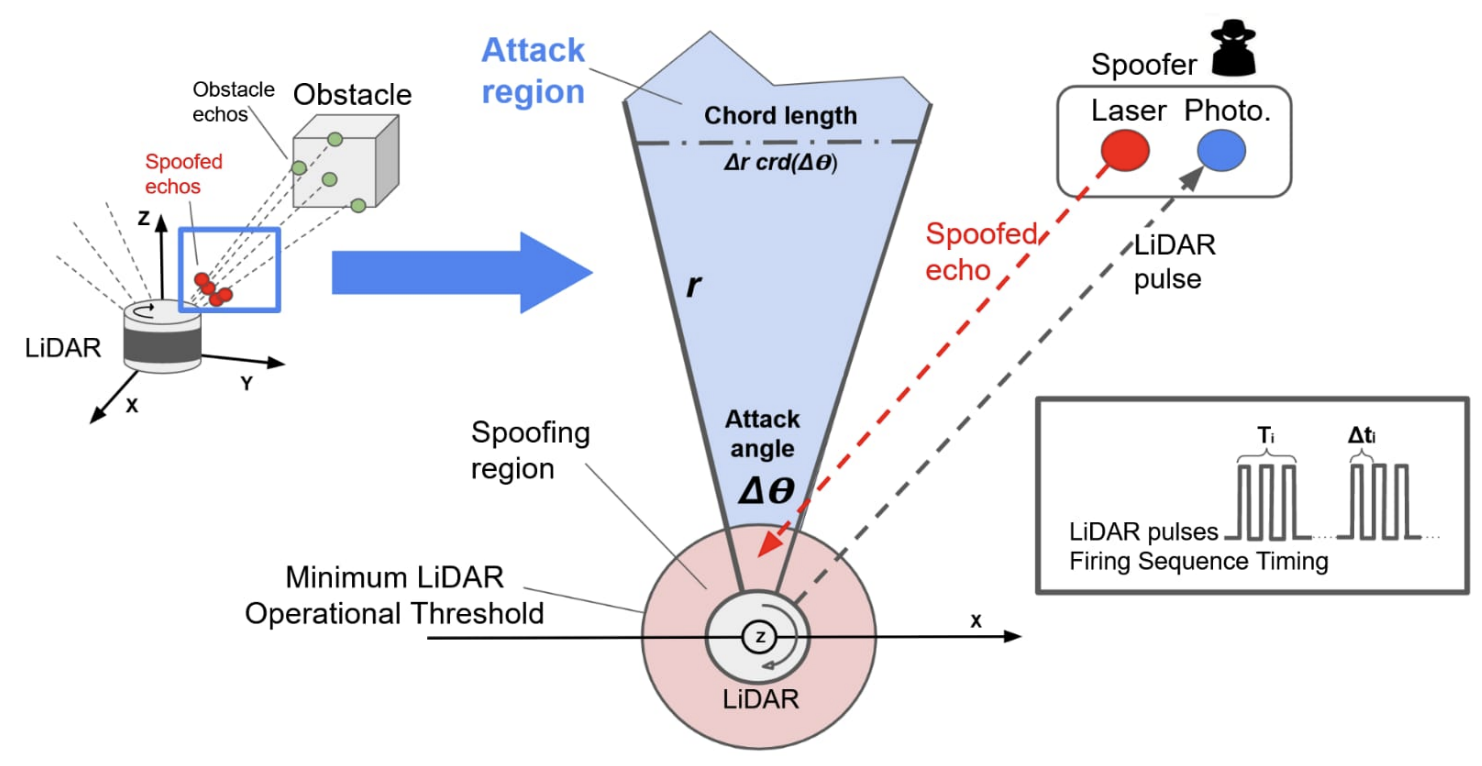}
    \caption{Overview of the Physical Removal Attack. The spoofer's photodiode receives laser pulses from the LiDAR and returns false echo signals that fall below the LiDAR's Minimum Operational Threshold (MOT), effectively removing objects from detection (Figure adapted from \cite{cao2023}).}
    \label{fig:cao2023}
\end{figure}

Cao et al. \cite{cao2023} introduce \textit{physical removal attack (PRA)}, which leverages laser spoofing to inject fake echoes close to the LiDAR sensor, causing it to ignore real obstacle points (see Figure \ref{fig:cao2023}). This technique exploits the inherent filtering and transformation processes of LiDAR data in AV systems. The paper demonstrates the effectiveness of PRA against three popular AV obstacle detectors (Apollo, Autoware, PointPillars) and evaluates the attack's impact on three fusion models (Frustum-ConvNet, AVOD, and Integrated-Semantic Level Fusion). The attack achieves a 92.7\% success rate in removing 90\% of a target obstacle's point cloud in moving vehicle scenarios.

The study includes empirical experiments using a Velodyne VLP-16 LiDAR sensor and evaluates the attack's impact on AVs in a production-grade simulator (LGSVL). It demonstrates that the attack can be executed even when vehicles are moving, using a tracking system to maintain synchronization with the target LiDAR sensor.

The authors show that existing defenses against LiDAR spoofing and object hiding attacks are ineffective against PRA. They propose two enhanced defense strategies, Fake Shadow Detection and Azimuth-based Detection, to mitigate this attack, achieving high true negative and true positive rates.

\begin{figure}
    \centering
    \includegraphics[width=1\linewidth]{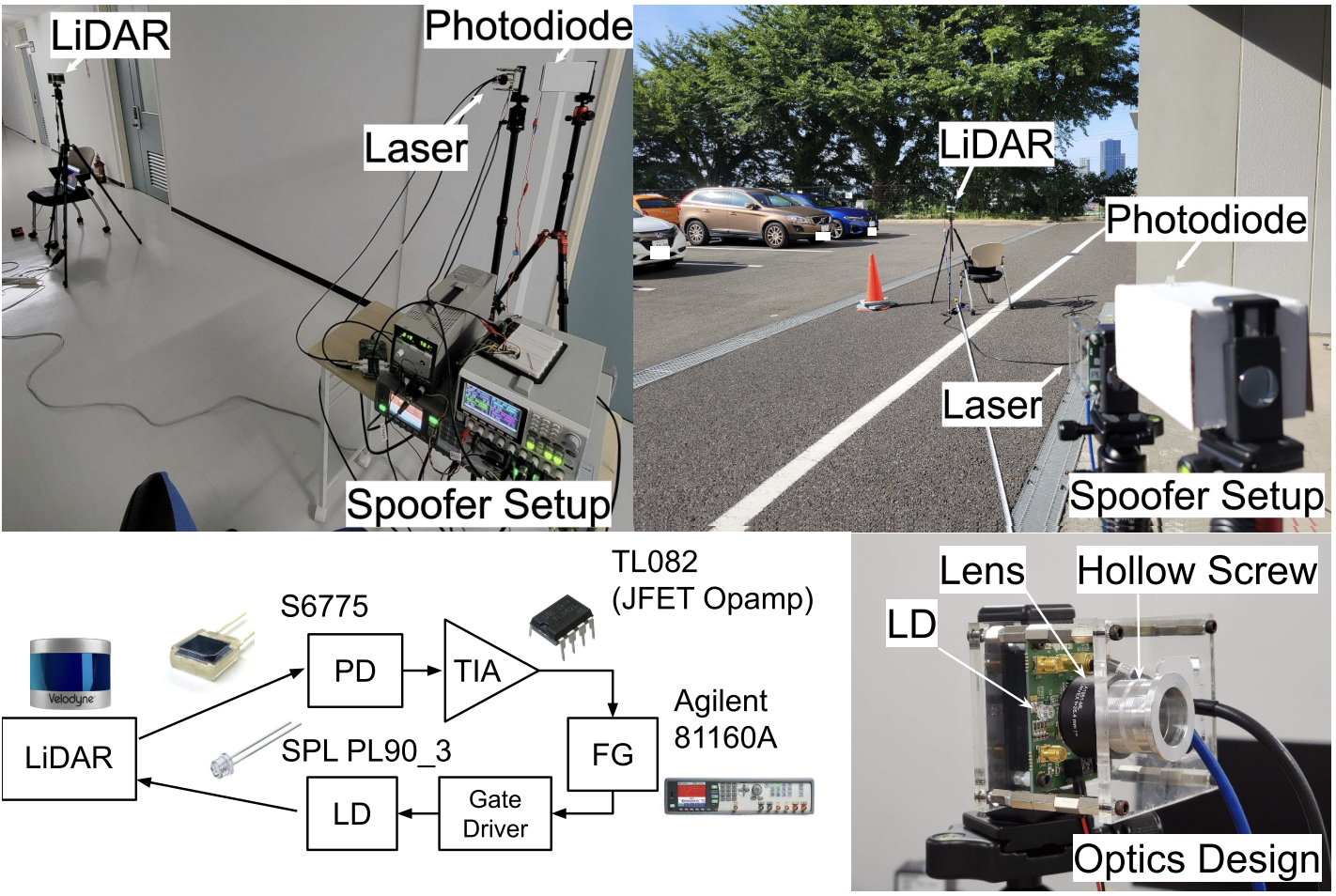}
    \caption{Overview of the LiDAR spoofer setup used in \cite{Sato2023}, including the optics design and the configurations for both indoor and outdoor experiments. Components include PD (Photodetector), TIA (Transimpedance Amplifier), FG (Function Generator), and LD (Laser Diode) (Figure adapted from \cite{Sato2023}).}
    \label{fig:sato2023}
\end{figure}

Sato et al. \cite{Sato2023} identify critical research gaps in prior LiDAR spoofing studies, such as focusing on a single LiDAR model (VLP-16), unvalidated attack capabilities, and limited evaluation of object detectors. The paper conducts the first large-scale measurement study on LiDAR spoofing attacks, covering 9 popular LiDAR models, including both first-generation (e.g., VLP-16) and new-generation LiDARs, and 3 major types of object detectors trained on 5 different datasets. It also introduces improvements in spoofing devices, as shown in Figure \ref{fig:sato2023}, significantly enhancing attack capabilities. The improved spoofer demonstrates the ability to inject over 6,000 spoofed points, surpassing previous methods which only managed up to 200 points.
\begin{figure}
    \centering
    \includegraphics[width=1\linewidth]{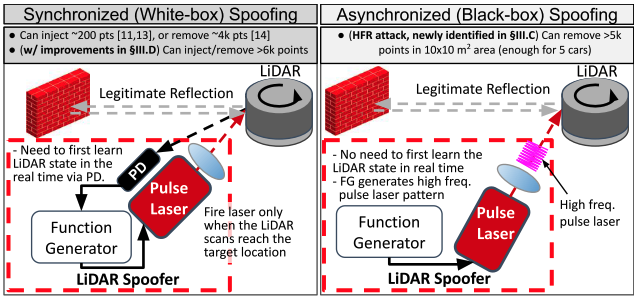}
    \caption{Illustration of the synchronized and asynchronized LiDAR spoofing techniques with the latest attack capabilities. (Figure adapted from \cite{Sato2023}).}
    \label{fig:sato2023sync}
\end{figure}

They introduce two new attack strategies (see Figure \ref{fig:sato2023sync}): 
\begin{itemize}
    \item \textbf{Object Removal Attack (ORA):} The paper identifies a new object removal attack for new-generation LiDARs, overcoming limitations of previous methods that required synchronization.
    \item \textbf{High-Frequency Removal (HFR) Attack:} It adapts the saturation attack to use high-frequency pulsed lasers instead of continuous lasers, achieving practical object removal without synchronization.
\end{itemize}

The study reveals that new-generation LiDARs, which feature timing randomization and pulse fingerprinting, exhibit different vulnerabilities to spoofing attacks compared to first-generation LiDARs. These security features challenge the feasibility of certain attack strategies, such as Chosen Pattern Injection (CPI).

Jin et al. \cite{jin2023} presents \textit{PLA-LiDAR}, a physical laser attack against LiDAR-based 3D object detection. The attack can inject adversarial point clouds into a LiDAR sensor with the correct shape and location, thereby hiding or creating objects. Authors developed a laser transceiver capable of injecting up to 4200 points into the LiDAR's perception, significantly increasing the number of spoofed points compared to prior work. 

They introduce four attack types:
\begin{itemize}
    \item \textbf{Naive Hiding:} Makes an existing object undetectable by creating a fake wall far away.
    \item \textbf{Record-based Creating:} Induces a non-existent object by injecting recorded point clouds.
    \item \textbf{Optimization-based Hiding:} Hides an existing object by injecting optimized adversarial points.
    \item \textbf{Optimization-based Creating:} Creates a non-existent object by injecting optimized adversarial points.
\end{itemize}

\begin{figure*}
    \centering
    \includegraphics[width=1\linewidth]{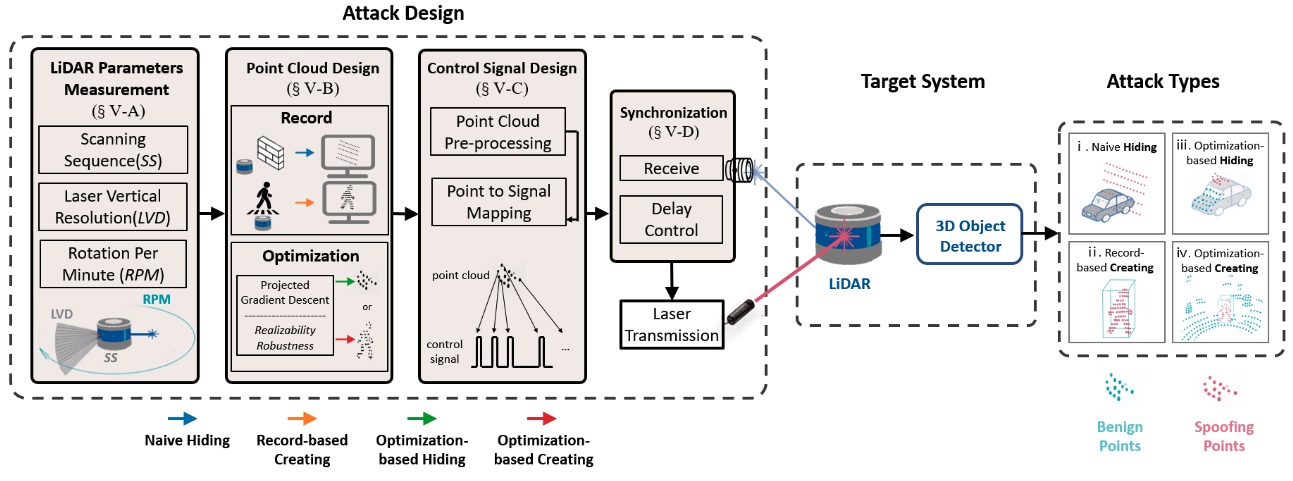}
    \caption{Attack workflow. The adversary begins by measuring the victim LiDAR and generates injectable point clouds either through recording or adversarial optimization. These point clouds are then converted into control signals, which are synchronized with the victim LiDAR to inject the lasers. This process can deceive the 3D object detector, leading to either object hiding or creation attacks (Figure adapted from \cite{jin2023}).}
    \label{fig:jin2023methodology}
\end{figure*}

The proposed methodology (see Figure \ref{fig:jin2023methodology}) is based on: First, measuring the LiDAR parameters; The attack targets mechanical LiDARs, which rotate and emit laser pulses to scan the environment. Key parameters include scanning sequence, laser vertical distribution, horizontal angular resolution, and wavelength. The attacker measures essential LiDAR parameters such as scanning sequence (to understand the firing and receiving order of laser pulses) and horizontal angular resolution. This involves using a substitute LiDAR of the same model as the victim's to capture these details accurately.
Second step, the design of Point Cloud; for Record-based attacks: the attacker uses a LiDAR of the same model as the target to record the point cloud of a real object, which can then be replayed to create a spoofed object in the target LiDAR's perception. This method doesn't require knowledge of the 3D object detection algorithms but is limited to scenarios where an exact object recording is feasible.

For optimization-based attacks, the attacker employs adversarial machine learning to generate point clouds with fewer points while considering the physical constraints of the LiDAR. Ensures generated points occur only on one of the LiDAR's laser rays, with each ray having at most one point. 
For hiding attacks, the method suppresses bounding box proposals near the target object. 
For creating attacks, it generates bounding boxes in the expected area to induce the perception of a non-existent object. Iteratively adjusts point positions in the spherical coordinate system to minimize the designed loss functions, considering the physical limits of the LiDAR system.

The next step is to ensure control signal design, the generated point cloud are converted into a series of laser pulses using a laser diode. The Point-to-Signal Mapping transforms the coordinates of each spoofed point into the timing of laser pulses.
The Timestamp Calculation determines when each laser pulse should be emitted to align with the scanning sequence of the target LiDAR. Control Signal Generation designs a control signal for a laser diode driver to emit these pulses precisely.

For the Synchronization, Aligning with LiDAR's Scanning Sequence synchronizes the attack signal with the victim LiDAR’s scanning sequence.
Uses a photodiode to detect when the LiDAR emits a laser pulse, generating a trigger signal that informs when to inject the spoofed laser pulses.
Delay Control introduces a precise delay in the laser emission to align the spoofed points with the target LiDAR’s scanning sequence.

\begin{figure*}
    \centering
    \includegraphics[width=1\linewidth]{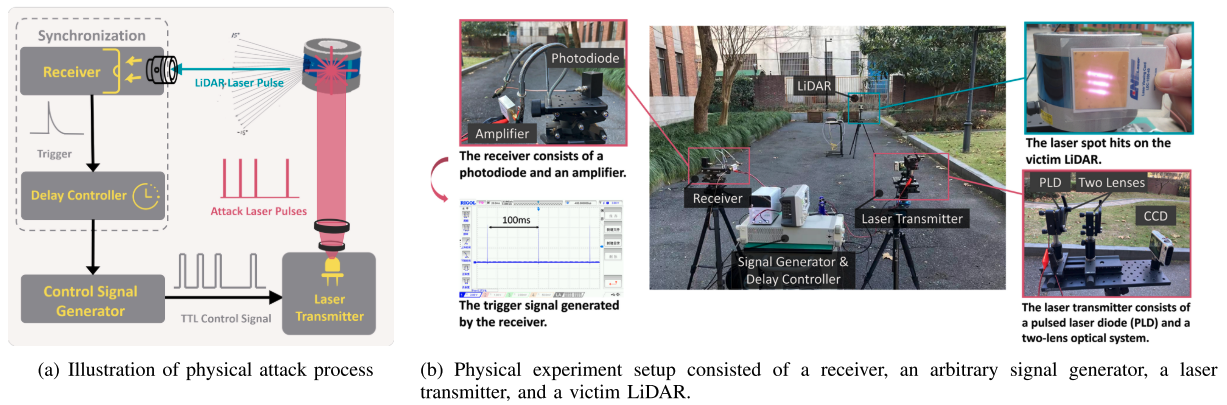}
    \caption{Illustration of the attack process and experimental setup for physical attacks (Figure adapted from \cite{jin2023}).}
    \label{fig:jin2023setup}
\end{figure*}

For the  attack device implementation, as shown in Figure \ref{fig:jin2023setup}, the attack system consists of:
\begin{itemize}
    \item A receiver (photodiode) to detect the victim LiDAR’s laser pulses.
    \item A delay controller and arbitrary waveform generator (AWG) to time the spoofing signals.
    \item A laser transmitter (laser diode and driver board) to emit the attack laser pulses.
\end{itemize}
This setup enables the attacker to inject thousands of spoofing points into the LiDAR’s perception system.



\begin{figure*}
    \centering
    \includegraphics[width=1\linewidth]{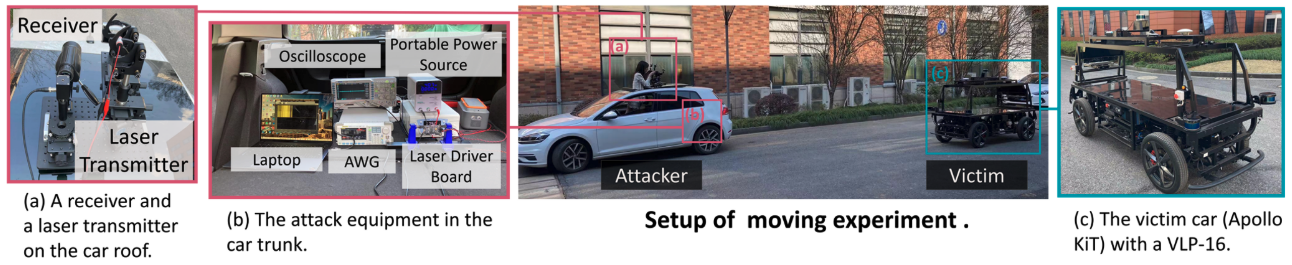}
    \caption{Experimental setup for physical attacks on moving vehicles (Figure adapted from \cite{jin2023}).}
    \label{fig:jin2023movingcar}
\end{figure*}
For the experimental setup for physical attacks on moving vehicles, as shown in Figure \ref{fig:jin2023movingcar}, it involves creating a real-world scenario where both the attacker and victim vehicles are in motion. The goal is to test the feasibility of PLA-LiDAR attacks on moving vehicles, simulating a situation where an autonomous vehicle (the victim) is being targeted by an attacker vehicle. Both the attacker and victim vehicles move at similar speeds to evaluate if the attack can still successfully deceive the LiDAR-based perception system under dynamic conditions.

The attacker Vehicle was equipped with the complete attack setup, including
a Laser Transmitter and Receiver mounted on the roof of the attacker vehicle. The receiver is connected to a gimbal for manual aiming to ensure it stays aligned with the victim LiDAR. For the Control Equipment, The arbitrary waveform generator (AWG), laser driver board, and other control devices are placed in the trunk of the vehicle. These components generate and control the timing of the laser pulses. The Power Source provides the necessary power for the laser transmitter and other electronic components.

The victim Vehicle was equipped with a LiDAR sensor, specifically a Velodyne VLP-16, which is commonly used in autonomous vehicles. The victim car is also integrated with a real-time 3D object detection system that uses the LiDAR data to perceive the environment.

For the driving setup, both vehicles drive at a slow speed of around 5 km/h for safety during the experiment. The attacker vehicle follows the victim vehicle at varying distances ranging from 5 to 15 meters to simulate different attack ranges. The attack system continuously emits laser pulses towards the victim LiDAR to inject spoofed point clouds. Manual aiming is done using the gimbal-mounted laser transmitter to maintain the alignment with the moving target LiDAR. The laser is adjusted to keep a steady spot on the victim LiDAR despite vehicle motion. The receiver uses a large-diameter telescope (50 mm) to expand its receiving area, improving the system's tolerance to minor misalignments caused by vehicle motion. The laser transmitter has an expanded spot diameter of 8 cm and employs a high-power laser diode (peak power of 300 W) to ensure adequate power intensity for effective point injection at varying distances.

The attacks showed high success rates even when the vehicles were moving. Specifically, hiding attacks achieved a 94.1\% success rate, and creating attacks had a 78.9\% success rate.

\begin{figure*}
    \centering
    \includegraphics[width=0.9\linewidth]{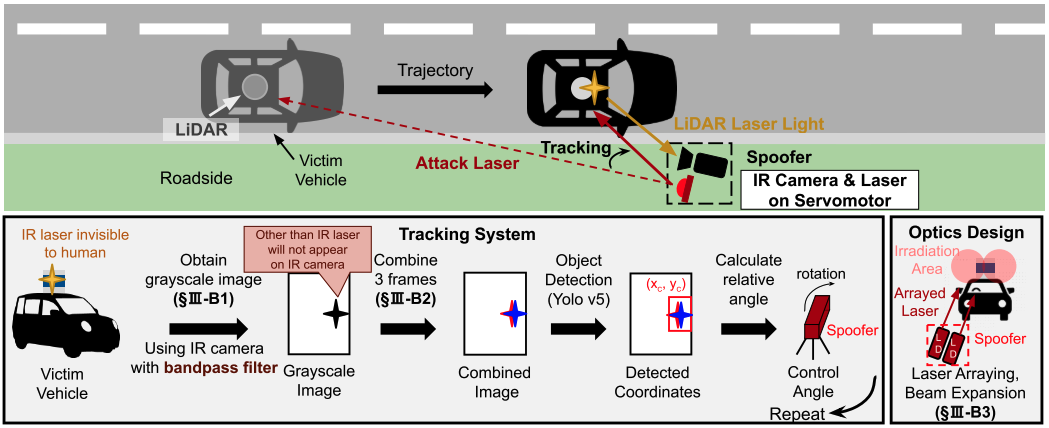}
    \caption{Overview of the MVS system and enhancements in optics design. By utilizing an IR camera and combining consecutive frames along the channel dimension, the system achieves stable tracking of a LiDAR at long distances. Additionally, the stability of laser irradiation is improved by arraying multiple attack lasers \cite{suzuki2024}).}
    \label{fig:suzuki2024}
\end{figure*}

Suzuki et al. \cite{suzuki2024} introduced a novel attack system called the \textit{Moving Vehicle Spoofing (MVS)} system, specifically designed to target vehicles traveling at high speeds. As illustrated in Figure \ref{fig:suzuki2024}, the MVS system leverages an infrared (IR) camera-based detection and tracking mechanism, combined with a precision aiming system. This aiming system is equipped with a high-precision servo motor and an array of laser diodes, enabling it to accurately spoof LiDAR sensors of moving vehicles. By synchronizing the movement of the attack system with the target vehicle, the MVS system can dynamically project adversarial signals that effectively deceive the vehicle's perception systems, creating significant risks in high-speed driving scenarios.

Zhu et al. \cite{zhu2024} propose a novel method called \textit{AE-Morpher} to enhance the robustness of adversarial attacks against LiDAR-based detection models. The key focus is on improving the physical-world effectiveness of adversarial objects by minimizing the discrepancies between the desired adversarial point cloud and the actual point cloud captured by LiDAR sensors. The proposed method focuses on reconstructing the adversarial object by identifying effective perturbations and expanding them to construct surfaces that are more likely to be captured by LiDAR sensors. This approach aims to optimize the presentation of adversarial perturbations to minimize distortions during the LiDAR capturing process, without interfering with the adversarial optimization stage. The approach increased the attack success rate (ASR) by an average of 38.64\% and reduced the adversarial ornament’s projection area by 67.59\%, making the attacks more efficient and harder to detect. Unlike previous methods requiring 3D printers, the adversarial objects reconstructed by AE-Morpher can be easily constructed by hand using low-cost materials like cardboard or wood board.

Kobayashi et al. \cite{Kobayashi2024} introduce a novel attack method. This method generates "\textit{Adversarial Shadows}" on the LiDAR point cloud. By placing materials like aluminum sheets strategically, the attacker creates shadows that mislead the object detection system, causing it to perceive non-existent objects. The attack takes advantage of the shadows naturally formed in the point cloud data captured by LiDAR sensors. By creating artificial shadows, the method fools object detection systems into false detections. The study presents scenarios where the attack can induce sudden stops in clear visibility or cause false evasive actions on multi-lane roads, leading to potential congestion or accidents. The attack involves a three-step process—acquiring point cloud data, optimizing the adversarial shadow, and implementing it using materials undetectable by LiDAR. The shadow’s location and angle are optimized to maximize the likelihood of false detections. Using the AWSIM simulator, the study evaluates the attack's effectiveness against two models, PointPillars (voxel-based) and Point-RCNN (point-based). The attack demonstrated a 100\% success rate against PointPillars in a flat environment and an average of 58\% in urban scenes.

\begin{table*}[!htp]
\centering
  \caption{Comparison of attack methods against only LIDAR-based perception. O – On the target object(s); P - In physical proximity.} 
  \label{tab:attacks_comparison}
  \begin{tabular}{|l|c|c|c|c|c|}
    \hline
       \textbf{Attack}     & \textbf{Placement} & \textbf{Test in physical domain}& \textbf{Practicality in road}  & \textbf{Test transferability} & \textbf{Evaluation metrics}\\
    \hline 
     Cao et al.\cite{cao2019ccs}           &  P & $\times$ &$\times$  &    $\times$ & Attack Success Rate\\ 
    \hline
     Cao et al. \cite{cao2019adversarial}  &  P & \checkmark &\checkmark  &   $\times$ & Attack Success Rate\\
    \hline
     Sun et al. \cite{sun2020}             & P & $\times$ &$\times$    &    $\times$ & Attack success rate  \\
    \hline
     Tu et al. \cite{tu2020}               &  O & $\times$&\checkmark  &    \checkmark & Attack success rate  \\
                                            &  &     &            &          & Recall-IoU curve     \\
    \hline
     Yang et al. \cite{yang2021}            & P & \checkmark    &  \checkmark  &   $\times$   & Detection Rate \\
     &  &   &      &    & Mis-classification Rate\\
     \hline
    Zhu et al. \cite{zhu2021}             & P & \checkmark    &    \checkmark    &   $\times$       &   Attack success rate   \\                        
    \hline
     Wang et al. \cite{wang2023}           & P & $\times$ &$\times$  &     $\times$  & Attack success rate \\ 
     \hline
     Sato et al. \cite{Sato2023}           & P & \checkmark &\checkmark  &     $\times$ & Attack Success Rate \\ 
                                     &     &     &    &   &Collision Rate\\ 
    \hline
     Cao et al. \cite{cao2023}       &  P   &  \checkmark   &  \checkmark   & $\times$ & Confidence Score \\ 
    \hline
     Jin et al. \cite{jin2023}  &   P   &  \checkmark   & \checkmark    & $\times$ & Attack Success Rate \\ 
  \hline
     Suzuki et al. \cite{suzuki2024}  &   P  &  \checkmark    &  \checkmark    & $\times$ & Attack Success Rate\\ 
  \hline
   Zhu et al. \cite{zhu2024}      &   O  &  \checkmark   &  \checkmark   & $\times$  & Attack Success Rate\\  
  \hline
   Kobayashi et al. \cite{Kobayashi2024}   &  P   & \checkmark &  \checkmark   & $\times$ & Attack Success Rate\\     
  \hline
\end{tabular} 
\end{table*}

\begin{table*}[!htp]
\centering
  \caption{Physical Spoofing Attacks on only LiDAR-based perception systems.}
  \label{tab:attacks_spoofing}
  \begin{tabular}{|l|c|c|c|c|c|c|c|}
    \hline
     \multirow{3}{*}{Attack}    & \multirow{3}{*}{LiDAR} & \multicolumn{4}{|c|}{Attack Capability} & Consider & multi-\\ \cline{3-6}
                &   &Number of   & Distance & Patterns & Shape  & Moving  & LiDAR\\    
              &  &Spoofing Points  & Control & Control & Control &  Vehicle  & Compatibility\\
    \hline
      Cao et al. \cite{cao2019ccs}   & VLP-16  & $\sim100pt$ & \checkmark & $\times$ &  $\times$ &  $\times$ &   $\times$ \\ 
    \hline
      Sun et al. \cite{sun2020}  & VLP-16  & $\sim200pt$ & \checkmark & $\times$ &  $\times$ &  $\times$ &     $\times$ \\  
    \hline
      Wang et al. \cite{wang2023} & VLP-16  & $\sim200pt$ & \checkmark & $\times$ & $\times$ &  $\times$ &     $\times$ \\  
    \hline
      Sato et al. \cite{Sato2023}  & VLP-16, 32c, XT32, Helios   & $> 6,000pt$ &  \checkmark & \checkmark  & \checkmark & $\times$ &    \checkmark \\  
    \hline
      Jin et al. \cite{jin2023}   & VLP-16, RS-16  & $\sim4200pt$ & \checkmark & \checkmark & \checkmark & \checkmark &   \checkmark \\  
    \hline

\end{tabular} 
\end{table*}

Table \ref{tab:attacks_comparison} presents a comparison of various attack methods targeting only LiDAR-based perception systems. It examines key aspects such as attack placement, whether the attacks were tested in the physical domain, their practicality and transferability, and the evaluation metrics used to assess their effectiveness.

Table \ref{tab:attacks_spoofing} provides an overview of various physical spoofing attacks targeting LiDAR-based perception systems, categorized by different attack capabilities and LiDAR models. The number of spoofed points is a crucial factor in the success of the attack. Higher numbers of spoofing points generally correlate with more effective deception of the LiDAR system. Cao et al. \cite{cao2019ccs}, Sun et al. \cite{sun2020}, and Wang et al. \cite{wang2023} focus on relatively small numbers of spoofed points (around 100 to 200 points), which limits the complexity of the spoofed object or scenario. Sato et al. \cite{Sato2023} and Jin et al. \cite{jin2023} demonstrate significantly more complex attacks, with over 6,000 and 4,200 spoofed points, respectively, allowing for more detailed object manipulation and potentially more realistic deception.

Table \ref{tab:attack_info} provides an analysis of robustness considerations and the defenses discussed in relation to various attack methods. It highlights how each method addresses robustness and the specific defensive strategies employed to mitigate the impact of these attacks.

Table \ref{tab:attacks_dataset} offers an overview of the tested deep learning models, autonomous driving platforms, datasets, simulators, and LiDAR models used in the evaluation of various attack methods.

\begin{table*}[!htp]
\centering
  \caption{Adversarial attacks against only LiDAR-based perception: Robustness consideration and Discussed defenses.}
  \label{tab:attack_info}
  \begin{tabular}{|l|l|l|}
    \hline
       \textbf{Attack}    & \textbf{Robustness} & \textbf{Discussed defenses}    \\
    \hline 
       Cao et al.\cite{cao2019ccs}     & Variations in Point Budget,  & AV System-Level defenses, \\
                 &    Variations in Distance Intervals & Sensor-Level Defenses    \\
    \hline
       Cao et al. \cite{cao2019adversarial}  &  Consider physical transformations,  & None    \\
                   & such as Variation in Positions and Orientations  &     \\
    \hline
       Sun et al. \cite{sun2020}  &  Variations of attack traces,  & Randomization-based defenses,   \\
                          & Variations in model performance  &      Adversarial Training,    \\        
                   &     & CARLO, SVF    \\
    \hline
      Tu et al. \cite{tu2020}   &  None  & Data Augmentation,     \\
                                &    & Adversarial Training    \\
    \hline
      Yang et al. \cite{yang2021}   &  Indoor vs Outdoor & Outlier Removal with kNN Distance, 
      \\
                                 &   & Random Noise 
                                 \\
    \hline

    Zhu et al. \cite{zhu2021} & Adversarial Object Size,   & Detecting Abnormal Point Cloud Patterns,     \\
         & Object Location Errors &      \\
    \hline
    Wang et al. \cite{wang2023}  &  Variations in Distance Intervals,  & None   \\
                                 &  Variations in Point Budget  &     \\
    \hline
    Sato et al. \cite{Sato2023}  & None   & Timing Randomization,     \\
                   &     & Pulse Fingerprinting,    \\
                &     & Simultaneous Laser Firing     \\
    \hline
     Jin et al. \cite{jin2023}   & Variations in Distance Intervals,  & None    \\
                &  Variations in Shape and Patterns  &    \\
    \hline
    Cao et al. \cite{cao2023}        & Different Lighting Conditions,   & Fake Shadow Detection,    \\
                            & Removing a Moving Obstacle   & Azimuth-based Detection   \\
    \hline
    Suzuki et al. \cite{suzuki2024} & Variations in Distance Intervals,  & Increasing the number of LiDARs,    \\
                            & Different Vehicle Speeds   & Infrared detection to track distant LiDARs    \\
  \hline
  Zhu et al. \cite{zhu2024} & Different Angles,  &  None    \\
                             & Different Distances &      \\
   \hline
   Kobayashi et al. \cite{Kobayashi2024}    & Object Size, Object Location, &   None   \\   
                      & Driving Directions and Distances, &  \\
                     &    Effect of passing-by vehicles &  \\
  \hline
\end{tabular} 
\end{table*}

\begin{table*}[!htp]
\centering
  \caption{Attacks on only LiDAR-based perception: DL Models, AD platforms, Datasets, Simulators, and LiDAR model.}
  \label{tab:attacks_dataset}
  \begin{tabular}{|l|l|l|l|l|l|}
    \hline
       \textbf{Attack}   & \textbf{DL Models} & \textbf{AD Platforms}& \textbf{Dataset} &\textbf{Simulator}  &\textbf{LiDAR} \\
    \hline 
       Cao et al.\cite{cao2019ccs}   & - & Apollo 2.5   & - &  Sim-control& Velodyne HDL-64E S3 \\
       \hline
       Cao et al. \cite{cao2019adversarial} & - & Apollo   & - &  - & Velodyne HDL-64E \\
       \hline
       Sun et al. \cite{sun2020} & PointPillars, PointRCNN  & Apollo 5.0 & KITTI &  - & Velodyne VLP-16 PUCK \\
       \hline
       Tu et al. \cite{tu2020} &  PIXOR, PIXOR (density),  & - & KITTI  &  -  & Velodyne HDL-64E\\
        &   PointRCNN, PointPillar &  &   &    & \\
        \hline
       Yang et al. \cite{yang2021} &PointRCNN, PointPillar, PV-RCNN & Baidu Apollo & - & LGSVL &  Velodyne VLP-16\\
       \hline
        Zhu et al. \cite{zhu2021}  &  PIXOR, VoxelNet, PointPillars, F-PointNet & - & KITTI  &  -  & Ouster OS1-64\\
        \hline
       Wang et al. \cite{wang2023} & PV-RCNN, PointPillars, PointRCNN,  & - & KITTI & - & Velodyne VLP-16, 32, 64 \\
                                       & IA-SSD, Voxel RCNN, PDV & & & & \\ 
        \hline
       Sato et al. \cite{Sato2023} & PointPillars, PV-RCNN, & Apollo 7.0, & KITTI,  & LGSVL & VLP-16 \cite{vlp16}, VLP-32c \cite{vlp32c},  \\
                                   & SECOND, PartA2, 3DSSD & Waymo       & Lyft,  & & VLS-128 \cite{vls128}, Pixell \cite{pixell},  \\
                                   &  &   & nuScenes &  & Realsense L515 \cite{realsense},    \\
                                     &  &   &  &  &   Horizon \cite{horizon},  OS1-32 \cite{os1},  \\  
                                           &  &   &  &  &  XT32 \cite{xt32}, Helios 5515 \cite{helios}  \\  
    \hline
     Jin et al. \cite{jin2023} &PointPillars, SECOND  & Apollo r6.5  & KITTI & - &   VLP-16, RS-16, HDL64E   \\  
     \hline
     Cao et al. \cite{cao2023}   & PointPillars, AVOD, Frustum-ConvNet &  Apollo 5.0,  & KITTI & LGSVL &  VLP-16   \\                               &               &  Autoware       &  &  &    \\   
     \hline
     Suzuki et al. \cite{suzuki2024}  & - & -  & Original  & - & VLP-16, Livox Horizon    \\ 
                                     &  &   & dataset &  &    \\   
    \hline
      Zhu et al. \cite{zhu2024}  & PointPillars, PointRCNN & Apollo 7.0  & KITTI  & LGSVL &  RS-LiDAR-16   \\ 
    \hline    
    Chen et al. \cite{chen2024} & PIXOR, VoxelNet, PointPillars, F-PointNet & -  & KITTI & - & -   \\  
    \hline
      Kobayashi et al. \cite{Kobayashi2024}  & PointPillars, PointRCNN & Autoware & KITTI & AWSIM &  -  \\   
  \hline
\end{tabular} 
\end{table*}


\section{Attacks on Sensor Fusion-Based Perception}
\label{fusion_perception}
In this section, we will explore various attacks targeting sensor fusion-based models. Perception systems that rely on Camera, LiDAR, and even Radar are inherently vulnerable to adversarial attacks \cite{eykholt2018robust, cao2019ccs, guesmi2022adversarial}. These attacks exploit the weaknesses of individual sensors as well as the fusion process itself, potentially leading to significant errors in object detection and classification.

Table \ref{tab:attack_fusion} represents different attacks on multi-sensor fusion-based perception. The table includes spoofing attacks, physical adversarial objects and reflective objects.

\begin{table*}[!htp]
\centering
  \caption{Main adversarial attack methods on Multi-Sensor Fusion-based perception: Attack Setting, Attacker Knowledge, Attack Goal (OI-Object Injection; OR-Object Removal; T-Translation; MC-Miscategorization), Attack Scenario, Attack Form, Perception and Venue.}
  \label{tab:attack_fusion}
  \begin{tabular}{|l|c|c|c|c|c|c|c|}
    \hline
       \textbf{Attack}  & \textbf{Setting} &\textbf{Knowledge} & \textbf{Goal} & \textbf{Form} &\textbf{Perception} & \textbf{Venue}  \\
    \hline 
      Wang et al. \cite{wang2021} & Digital  & Black-box & OI  & Adversarial points  & Multi-Modality & ICICS 2021\\
      \hline
      Liu et al. \cite{liu2021} & Digital  & Black-box & OR  & Adversarial points & Multi-Modality & KDD 2021\\
      \hline
      Abdelfattah et al. \cite{abdelfattah2021} &  Physical & White-box & OR & Physical Adversarial Object & Multi-Modality  & ICIP 2021\\
      \hline
      Abdelfattah et al. \cite{Abdelfattah2021iros}&  Physical & White-box & OR & Physical Adversarial Object & Multi-Modality  & IROS 2021\\
      \hline
      Cao et al. \cite{cao2021} & Physical  & White-box &  OR  & Physical Adversarial Object & Multi-Modality & S\&P 2021\\
      \hline
      Hallyburton et al.\cite{hallyburton2022} & Physical & Black-box  & OI,OR,T & Spoofing Attack & Multi-Modality & USENIX 2022\\
      \hline
      Yang et al. \cite{yang2023} & Physical  & White-box  & OR & Spoofing Attack & Multi-Modality & EI2 2023\\
      \hline
      Zhu et al. \cite{zhu2024malicious} & Physical  & White-box & OR  & Reflective Objects & Multi-Modality & MobiCom 2024\\
 \hline
\end{tabular} 
\end{table*}


Wang et al. \cite{wang2021} explore the vulnerability of MSF models by attacking only the LiDAR channel to generate adversarial point clouds, proving that even MSF models can be fooled into detecting a fake near-front object with high confidence. In \cite{wang2021}, a black-box attack method based on a genetic algorithm is proposed. This method generates adversarial point clouds with few points without needing access to the specific structures and parameters of the models, making it simpler and transferable.

The authors analyze the robustness of the AVOD model against a general spoofing attack on LiDAR-based models and find that it is invalid due to the correction of the camera channel. They generate adversarial point clouds using a genetic algorithm to attack the LiDAR channel alone. The attack is evaluated on different combinations of points and distances, and universal adversarial examples are generated at the best distance.

The attack achieves a high success rate, more than 95\% on the KITTI validation set when using point clouds with more than 30 points at an optimal distance of 4 meters. It also achieves average confidence scores over 0.9. The study verifies the transferability of the generated universal adversarial point clouds across models, demonstrating the attack's generality.

Liu et al. \cite{liu2021} propose a novel black-box adversarial attack that exploits the correlations between camera images and LiDAR point cloud data to generate adversarial examples. It targets multi-sensor fusion models that combine data from both sensors to enhance robustness.
The approach starts by generating adversarial image examples using a Generative Adversarial Network (GAN) based framework and an auxiliary image semantic segmentation model. The perturbations from the adversarial images are mapped onto the LiDAR point cloud space using three different strategies:

\begin{itemize}
    \item \textbf{Location Mapping:} Uses the geometric configuration of the camera and LiDAR to project image perturbations onto the point cloud.
    \item \textbf{Linear Transformation:} Learns a transformation matrix to project perturbations from the image view to the point cloud view.
    \item \textbf{Canonical Correlation Analysis (CCA):} Uses CCA to maximize the correlation between image and point cloud data in a latent subspace for effective perturbation projection.
\end{itemize}

\begin{figure*}
    \centering
    \includegraphics[width=0.8\linewidth]{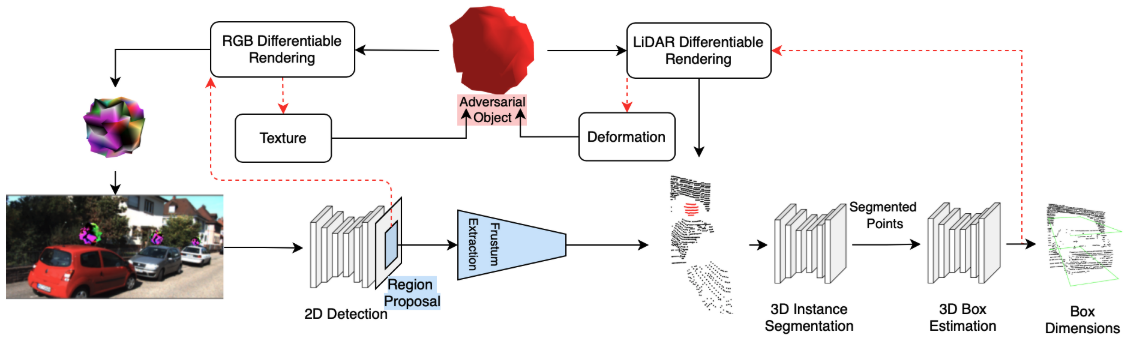}
    \caption{Overview of the attack pipeline (Figure adapted from \cite{abdelfattah2021}).}
    \label{fig:abdelfattah2021}
\end{figure*}
The proposed attacks, particularly using CCA-based projection, resulted in significant drops in detection performance of the fusion-based 3D detector (EPNet). The method was also tested against a LiDAR-only object detector, demonstrating its general applicability.

Abdelfattah et al. \cite{abdelfattah2021} propose an attack that uses a single 3D mesh with perturbable geometry and texture to deceive both the camera and LiDAR components of the detection model. This adversarial object is designed to be placed on top of a car, significantly reducing the model’s ability to detect the car. Unlike previous attacks that were either limited to the digital domain or not physically realizable, this method ensures the adversarial object can be created and positioned in the physical world, making it a practical threat to real-world systems.

As illustrated in Figure \ref{fig:abdelfattah2021}, the attack involves training a 3D mesh with learnable geometry and texture. The shape and color of the mesh are optimized to deceive the cascaded detection model when viewed from multiple angles. The paper uses differentiable rendering to simulate the adversarial object in both point clouds and RGB images, allowing the model to be deceived during inference. The attack aims to minimize the probability of correctly detecting the car by manipulating the object’s geometry and texture to generate adversarial perturbations.

In \cite{Abdelfattah2021iros} the same authors expanded and evaluated the previously introduced adversarial mesh object-based attack against both Frustum PointNet [118] and EPNet [136] which was a deep/intermediate fusion network. The two attack methods mentioned above were primarily designed to prevent the network from detecting actual objects that have the adversarial mesh on top of them. 

\begin{figure*}
    \centering
    \includegraphics[width=\linewidth]{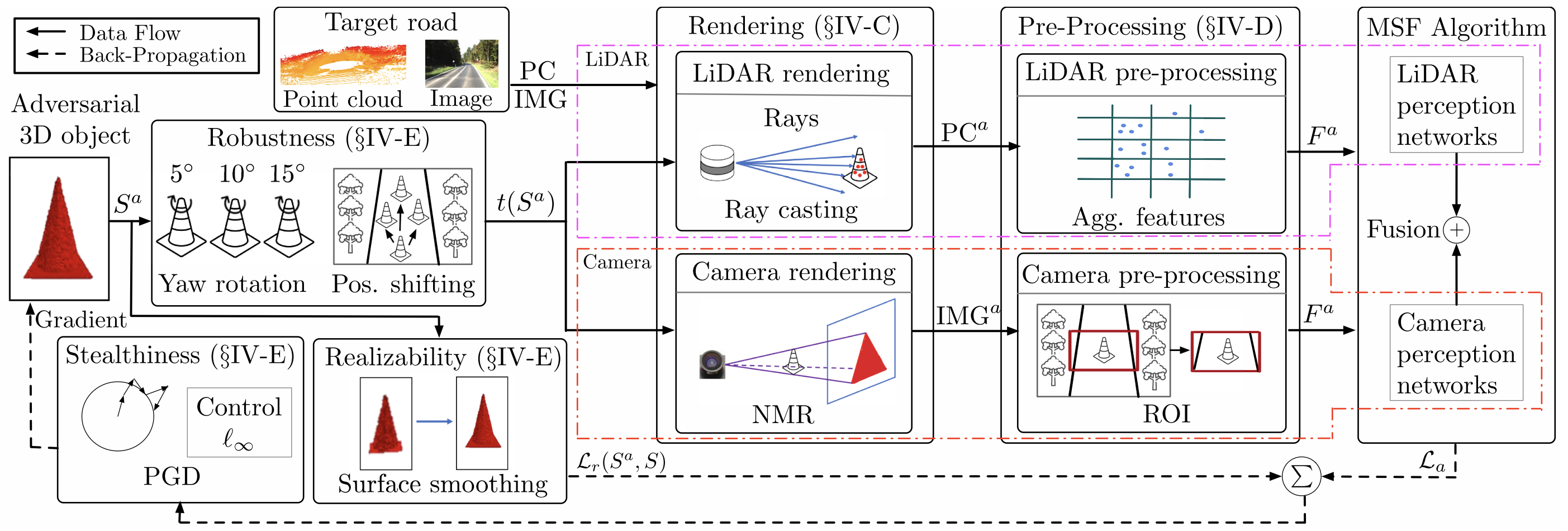}
    \caption{Overview of the optimization-based adversarial 3D object generation in MSF-ADV (Figure adapted from \cite{cao2021}).}
    \label{fig:CAO2021}
\end{figure*}
The proposed attack was tested on the KITTI dataset, a standard benchmark for 3D object detection in autonomous driving. The attack reduced the average precision of car detection by nearly 73\% under easy conditions, showing a significant impact on the detection capabilities of the target model. The study found that attacking both the camera and LiDAR modalities simultaneously was more effective than targeting either modality alone.

Cao et al. \cite{cao2021} propose a physically realizable adversarial attack using a 3D-printed object designed to mislead both camera and LiDAR sensors in MSF-based AD systems. The attack causes the vehicle to fail in detecting the object, potentially leading to collisions.
The proposed attack method, \textit{MSF-ADV}, generates an adversarial 3D object by manipulating its shape to deceive both camera images and LiDAR point clouds (see Figure \ref{fig:CAO2021}). The method addresses two main challenges:
\begin{itemize}
    \item Synthesizing physically consistent impacts on both camera and LiDAR.
    \item Handling non-differentiable cell-level aggregated features used in LiDAR perception.
\end{itemize}

\begin{figure}
    \centering
    \includegraphics[width=\linewidth]{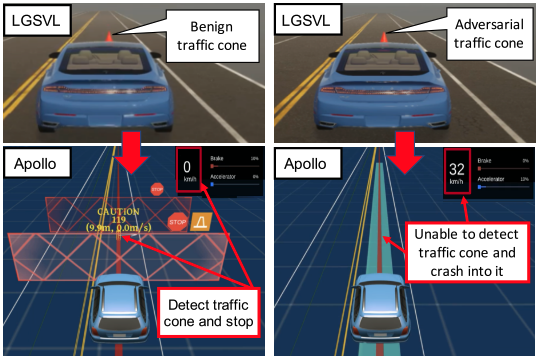}
    \caption{Screenshots of Apollo and LGSVL in the end-to-end attack evaluation with benign and adversarial traffic cones. Across 100 runs, the crash rate is 100\% for adversarial case, and 0\% for benign case (Figure adapted from \cite{cao2021}).}
    \label{fig:cao2021_lgsvl}
\end{figure}

The attack achieves a success rate of over 90\% across different object types and MSF algorithms in real-world driving scenarios. It demonstrates the ability to evade detection with high stealthiness and robustness to different vehicle positions (see Figure \ref{fig:cao2021_lgsvl}). The study evaluates the attack using 3D-printed objects and real LiDAR and camera devices, confirming the attack's effectiveness in the physical world. In simulated environments, the attack caused a 100\% vehicle collision rate for an industry-grade AD system. 
The paper also discusses potential defense strategies and evaluates existing ones, highlighting the need for improved security measures in MSF-based perception systems.

Hallyburton et al.\cite{hallyburton2022} introduce a novel attack method called the "\textit{frustum attack}", which targets camera-LiDAR fusion systems by maintaining consistency between camera and LiDAR data. This attack is context-aware and significantly compromises the perception algorithms of AVs, even those using multi-sensor fusion. The study evaluates eight widely used perception algorithms across three types of architectures (LiDAR-only and camera-LiDAR fusion). It demonstrates that all these algorithms are vulnerable to the frustum attack, showing that even fusion-based models are not immune to sophisticated spoofing attacks.

The frustum attack is shown to be stealthy, bypassing existing defenses against LiDAR spoofing as it preserves consistencies between camera and LiDAR semantics. The attack can consistently deceive the tracking modules in AV systems, creating adverse outcomes on end-to-end AV control. The frustum attack is capable of generating false positives (FPs) and false negatives (FNs), leading to dangerous driving behaviors such as sudden stops or false evasive actions.

The paper reveals that current defenses, such as CARLO \cite{sun2020}, SVF \cite{sun2020}, and ShadowCatcher \cite{hau2021shadow}, are not effective against the frustum attack. This highlights the need for new defense strategies to protect AVs from such sophisticated spoofing attacks.

\begin{figure*}
    \centering
    \includegraphics[width=0.8\linewidth]{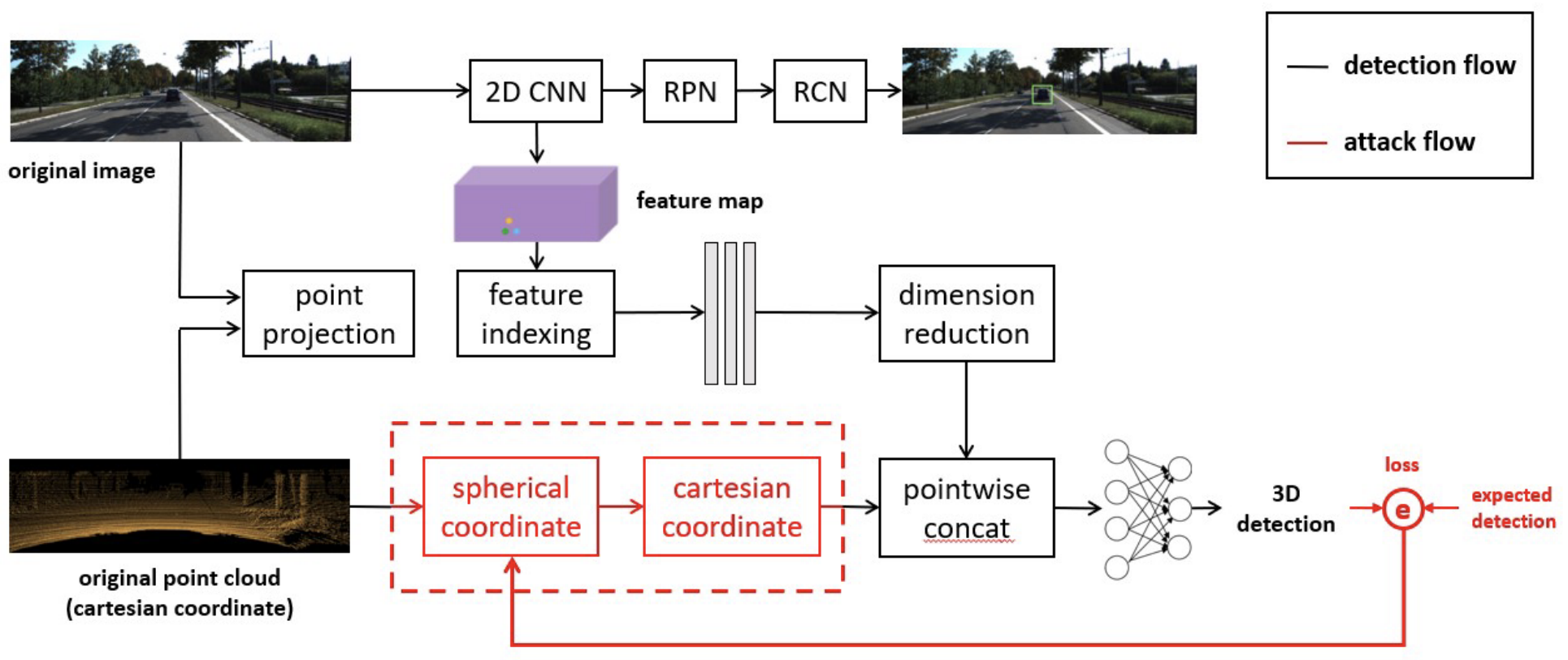}
    \caption{Overview of the attack pipeline. The standard data processing flow of MVX-Net is indicated by solid black lines. To ensure the injectability of adversarial points into the victim LiDAR in the real world, these points are transformed based on a specific angle in the spherical coordinate system, using the LiDAR as the reference point. This additional data processing step is denoted by a solid red line. The flow of gradients is shown by dashed red lines. As the entire pipeline is differentiable, gradients can flow from the adversarial loss back to the adversarial points within the spherical coordinate system. Ultimately, cars with adversarial points situated above their roofs are rendered undetectable by the fusion model (Figure adapted from \cite{yang2023}).}
    \label{fig:yang2023}
\end{figure*}

Yang et al. \cite{yang2023} propose a method that injects adversarial points into the LiDAR data channel (see Figure \ref{fig:yang2023}). These points are designed to evade detection while adhering to physical constraints, such as alignment with LiDAR rays and specific angular ranges, to ensure real-world feasibility. By targeting only the LiDAR data channel, the attack successfully deceives the fusion model without altering the image data channel, raising safety concerns for autonomous driving systems.

The study uses the MVX-Net fusion model and evaluates how the number of adversarial points, distance, and angle between the target and the LiDAR-equipped vehicle affect the attack success rate (ASR). Results show that increasing the number of adversarial points generally leads to a higher ASR, and cars farther away from the LiDAR sensor are easier to hide. 

Zhu et al. \cite{zhu2024malicious} proposed a method that employs a single type of adversarial object that passively reflects signals to fool all three sensor types (LiDAR, camera, and radar). This object can be easily fabricated at a low cost and used in practice with high stealthiness and flexibility. 

The adversarial object combines a smooth metal surface to deflect radar signals, a color patch to manipulate camera perception, and reflective properties to interfere with LiDAR detection. By placing these objects in specific locations with certain orientations, the attacker can hide a target vehicle from the AV’s multi-sensor fusion system.

The paper describes how attackers can use drones or other carriers to position the adversarial objects around a target vehicle, causing it to be hidden from the perception system of a victim AV. The attack can continuously hide a target vehicle from the perception system using only two small adversarial objects. Experiments on a real-world AV testbed show a high attack success rate. The paper also introduces a framework to analyze the vulnerability of sensor fusion systems, identifying systems that rely heavily on a subset of sensors and providing guidance for designing more robust fusion systems.

Table \ref{tab:fusion_comparison} presents a comparison of attack methods on multi-sensor fusion-based perception systems. It includes key aspects such as attack placement, whether the attacks were tested in the physical domain, their practicality for real-world road scenarios, the consideration of transferability, and the evaluation metrics used to assess their impact.

Table \ref{tab:fusion_info} outlines the robustness elements that were considered for each attack method on multi-sensor fusion-based perception. It also indicates whether any defense methods were discussed to counteract these attacks.

Table \ref{tab:fusion_dataset} illustrates the various deep learning models, autonomous driving platforms, datasets, simulators, and LiDAR models that were tested in the context of multi-sensor fusion-based perception attacks.

\begin{table*}[!htp]
\centering
  \caption{ Comparison of attack methods on Multi-Sensor Fusion-based perception. O – On the target object(s); P - In physical proximity.} 
  \label{tab:fusion_comparison}
  \begin{tabular}{|l|c|c|c|c|c|}
    \hline
       \textbf{Attack}     & \textbf{Placement} & \textbf{Test in physical domain} & \textbf{Practicality in road} & \textbf{Test transferability} & \textbf{Evaluation metrics}\\
    \hline 

     Wang et al. \cite{wang2021}           & P &  $\times$  &  $\times$      & \checkmark &  Attack success rate \\
                                               &  &    &     &   & Average Confidence Score (ACS)   \\
    \hline
     Liu et al. \cite{liu2021}               & P &  $\times$  &  $\times$   & \checkmark  &  Average precision \\
     \hline
     Abdelfattah et al. \cite{abdelfattah2021}   & O & $\times$  &   \checkmark   &  $\times$  & Average precision\\
          \hline
     Abdelfattah et al. \cite{Abdelfattah2021iros}   & O & $\times$  &   \checkmark   &  $\times$  & Average precision\\
     \hline
     Cao et al. \cite{cao2021}              & P & \checkmark   &   \checkmark  & \checkmark   & Attack success rate \\    
     \hline
     Hallyburton et al.\cite{hallyburton2022} & O,P &  \checkmark  &  \checkmark   &  $\times$   & Attack success rate \\  
     \hline
     Yang et al. \cite{yang2023}           & P &  $\times$ &  \checkmark  & $\times$  & Attack success rate\\ 
     \hline
     Zhu et al. \cite{zhu2024malicious}    & O,P  & \checkmark &  \checkmark  &    $\times$    & Detection Recall\\  
  \hline
\end{tabular} 
\end{table*}

\begin{table*}[!htp]
\centering
  \caption{Adversarial attacks on Multi-Sensor Fusion-based perception: Attacker's knowledge, Robustness consideration and Discussed defenses.}
  \label{tab:fusion_info}
  \begin{tabular}{|l|l|l|l|}
    \hline
       \textbf{Attack}   & \textbf{Robustness Consideration} & \textbf{Discussed defenses}   \\
    \hline 
      Wang et al. \cite{wang2021}  & Variations in Point Budget, & None 
      \\
                    & Variations in Distance Intervals  &   \\
    \hline
      Liu et al. \cite{liu2021}    & Impact of the Selection Ratio  & Adversarial Training 
      \\
    \hline
      Abdelfattah et al. \cite{abdelfattah2021} & Variation in View Point  & Adversarial Training 
      \\
        \hline
      Abdelfattah et al. \cite{Abdelfattah2021iros} & Variation in View Point  & Adversarial Training 
      \\
        
    \hline
     Cao et al. \cite{cao2021} & Variations in Distance Intervals  &  Input Transformation  
     \\
                    &   &   Adversarial Training  
                    \\
                     &   &   Certified Robustness 
                     \\
    \hline
    Hallyburton et al.\cite{hallyburton2022}  & Variations in Distance Intervals  & CARLO, SVF, ShadowCatcher, LIFE    \\
    \hline
    Yang et al. \cite{yang2023}  & Variation in Angle  & None 
    \\
                     & Variation in Distance  &     \\
     \hline
    Zhu et al. \cite{zhu2024malicious}  & Drones Stability, Vehicle Speed,  & Adversarial Training      \\
              &    Vehicle Direction, Variation in Distance &     \\
  \hline
\end{tabular} 
\end{table*}

\begin{table*}[!htp]
\centering
  \caption{Attacks on Multi-Sensor Fusion-based perception: Models, Datasets, Simulators, and Code.}
  \label{tab:fusion_dataset}
  \begin{tabular}{|l|l|l|l|l|l|}
    \hline
       \textbf{Attack}   & \textbf{DL Models} & \textbf{AD Platforms}& \textbf{Dataset} &\textbf{Simulator}  & \textbf{LiDAR model} \\
    \hline 
       Wang et al. \cite{wang2021} & AVOD, EPNet & - & KITTI &  -  & -\\
       \hline
       Liu et al. \cite{liu2021} & EPNet, PointRCNN & - &  KITTI &   -  & -\\
       \hline
       Abdelfattah et al. \cite{abdelfattah2021} & Frustum-PointNet (F-PN) $+$ YOLOv3 & - & KITTI & - & \\
       \hline
       Abdelfattah et al. \cite{Abdelfattah2021iros} & Frustum-PointNet (F-PN), EPNet  $+$ YOLOv3 & - & KITTI & - & \\
       \hline
       Cao et al. \cite{cao2021} & YOLOv3 & Baidu Apollo, Autoware & KITTI &   LGSVL & Velodyne HDL-64E \\
       \hline
       Hallyburton et al.\cite{hallyburton2022} & PointPillars, PointRCNN, BEV-based PIXOR &Baidu Apollo & KITTI & LGSVL & - \\
       \hline
       Yang et al. \cite{yang2023} & MVX-Net & - & KITTI & - & Velodyne HDL-64E \\
       \hline
       Zhu et al. \cite{zhu2024malicious} & BEVFusion, CRFNet, Radarnet, LFusion & Baidu Apollo & KITTI & - & Velodyne VLP-32C\\
             & RRPN, HD-FPNet, YOLOv3  &  &  &  & \\
  \hline
\end{tabular} 
\end{table*}

\section{Defenses}
\label{defenses}
To counteract LiDAR spoofing attacks, several defense mechanisms have been proposed. These include model-agnostic defenses that operate independently of the perception model, such as CARLO \cite{sun2020}, ShadowCatcher \cite{hau2021shadow}, Shadow-based Detection (Hau et al. \cite{hau2022defense}), and FDII \cite{zhang2023cooperative}, as well as model-based defenses that aim to enhance the perception architecture itself, like SVF \cite{sun2020} and LIFE \cite{liu2021seeing}.

CARLO \cite{sun2020} focuses on detection and is designed to protect LiDAR-only perception systems from basic spoofing attacks, particularly in near-front positions. It operates on the principle that if numerous LiDAR points seem to pass through a detected object, that object is likely to be a false positive (FP).

ShadowCatcher \cite{hau2021shadow}, another detection-centric defense, utilizes a similar principle to CARLO. It identifies objects as potential false positives if they possess highly anomalous shadow regions, determined by a high anomaly score based on the features of the shadow region.

Hau et al. \cite{hau2022defense} leverages the physical phenomenon of 3D shadows in LiDAR point clouds to detect hidden objects that evade conventional object detectors. The key idea is that while adversarial objects may be hidden from 3D detectors, they still occlude LiDAR pulses, creating detectable shadow artifacts in the point cloud. This methodology provides an orthogonal defense against object hiding attacks by using a physical property (shadows) rather than relying solely on DNN-based detectors.

Zhang et al. \cite{zhang2023cooperative} propose a defense (FDII) that leverages cooperative perception, where multiple vehicles share LiDAR scan data to identify discrepancies caused by spoofing attacks. The defense relies on exchanging LiDAR scan data among nearby vehicles. Since spoofing attacks typically target only one vehicle at a time, the shared data from unaffected neighboring vehicles can be used to detect anomalies.

SVF \cite{sun2020} is a model-based defense that aims to safeguard LiDAR perception by adding a point-wise confidence score to the LiDAR data in the front-view (FV). The underlying intuition is that naive false positives do not maintain consistency in the front view.

LIFE \cite{liu2021seeing} takes a hybrid approach by integrating LiDAR and camera data to provide a more robust defense against spoofing. This method cross-checks sensor detections using object matching between camera and LiDAR data in the front view. It also compares raw sensor data by evaluating the consistency of camera feature points with the LiDAR data in a depth image. Additionally, it uses machine learning algorithms to assess the reliability of sensor data by comparing predicted sensor values with actual captured data. LIFE was tested against basic spoofing attacks using LiDAR and stereo imagery, but it did not analyze the full spectrum of spoofing performance.

The integration of multi-sensor fusion has been suggested as a way to improve perception resilience \cite{cao2019ccs, sun2020, liu2021seeing, pajic2014robustness, ivanov2014attack}. However, there has been no comprehensive evaluation of how sensor fusion performs under spoofing attacks. For instance, while LIFE \cite{liu2021seeing} was evaluated using simple spoofing attacks, it did not delve into a detailed analysis of spoofing performance. Similarly, \cite{cao2021} used optimized physical adversarial objects as the threat model but did not conduct a systematic evaluation of sensor fusion.

\section{Discussion}
\label{discussion}



\subsection{Impact on AV Perception}
Adversarial attacks on LiDAR systems can severely compromise the perception capabilities of autonomous vehicles, leading to significant safety and operational risks. The key impacts include:

\subsubsection{Degradation of Detection Accuracy} Adversarial attacks can drastically impair the accuracy of object detection by distorting the LiDAR point cloud data. This manipulation confuses perception algorithms, preventing them from correctly identifying and classifying objects. As a result, critical elements like pedestrians, other vehicles, or road signs may be misidentified or completely missed. Such inconsistencies in detection can lead to unreliable perception, where objects are only intermittently recognized, increasing the likelihood of accidents. Furthermore, these attacks can compromise distance measurement accuracy, making it difficult for the vehicle to gauge the proximity of obstacles accurately, which is essential for safe navigation and decision-making.

\subsubsection{Inducing False Positives and Negatives} These attacks can result in the system generating false positives (detecting non-existent objects) or false negatives (failing to detect actual objects), both of which pose serious risks in autonomous driving. False positives can lead to the detection of phantom obstacles, causing unnecessary braking, swerving, or other evasive maneuvers that may disrupt traffic flow or even lead to accidents. On the other hand, false negatives are equally dangerous, as they may cause the vehicle to overlook real obstacles, such as pedestrians or other vehicles, increasing the risk of collisions and creating hazardous road conditions.

\subsubsection{Elevated Navigation and Safety Risks} The degradation of perception accuracy and the introduction of false readings significantly elevate navigation and safety risks for autonomous vehicles. Inaccurate perception data can lead to poor navigation decisions, resulting in erratic or unpredictable driving behavior. This increases the likelihood of accidents, as the vehicle may either fail to respond appropriately to actual hazards or overreact to non-existent threats. Persistent misinterpretations and unsafe maneuvers caused by compromised LiDAR perception not only pose direct safety risks but also erode public trust in autonomous vehicle technology, hindering its broader adoption.

\subsection{Affected Downstream Tasks.}
Conceptually, any downstream task that uses LiDAR point cloud data as input could be susceptible to the attacks presented in this paper. This includes geometric vision tasks like registration, pose estimation, and mapping, as well as pattern recognition tasks such as 3D object detection \cite{lang2019pointpillars, liang2019multi, shi2019pointrcnn, wang2020pillar, yang20203dssd, zhou2018voxelnet}, semantic segmentation \cite{hu2020randla, jiang2020pointgroup}, motion prediction \cite{luo2018fast, wu2020motionnet, zeng2019end}, and multiple object tracking \cite{weng20203d, weng2020gnn3dmot}. 
While the first set of tasks may be less impacted due to potential data fusion with GNSS signals, the latter set is more vulnerable. The reason is that small, carefully crafted perturbations in the point cloud data can significantly disrupt the functioning of deep learning models, as demonstrated in previous studies. This underlines the heightened risk to tasks dependent on accurate and reliable 3D perception.

\subsection{Open Research Challenges and Future Trends}

\noindent\textbf{Transferability:}  
One of the primary challenges in adversarial attacks on LiDAR-based perception systems is enhancing the transferability of these attacks, particularly in black-box scenarios. Future research should focus on developing techniques that enable attacks to be successfully transferred to models with limited access, where direct interaction or querying of the target models is not feasible. This requires creating more universal perturbations capable of effectively deceiving various architectures and operating across different scenes, without relying on specific model details.

\noindent\textbf{Adapting to New-Generation LiDAR Systems:}
With the advent of new-generation LiDAR systems featuring enhanced resolution, range, and noise resistance, it is essential to understand how these advances affect both the efficacy of adversarial attacks and the design of defense mechanisms. Future research should explore how these new LiDAR technologies can be leveraged to improve robustness against attacks while also investigating potential vulnerabilities introduced by these advanced capabilities. Understanding the interplay between emerging LiDAR technology and adversarial techniques will be key to developing more secure autonomous systems.

\noindent\textbf{Enhancing Robustness in Real Road Environments:}  
Current adversarial object generation methods often fail to account for real-world physical characteristics, leading to a reduced success rate when deployed in outdoor driving conditions. Future research should aim to refine simulation tools and attack processes to incorporate factors such as changing lighting conditions, sensor noise, and dynamic environmental elements. By better simulating these real-world complexities, we can develop adversarial attacks that are not only effective in controlled environments but also maintain their potency in varied and unpredictable real-world settings.

\noindent\textbf{Developing Advanced Defense Mechanisms:}  
As adversarial attacks continue to evolve, there is an urgent need to explore and develop more sophisticated defense mechanisms for LiDAR-based perception systems. Future research should focus on enhancing existing defenses and creating innovative strategies that can effectively mitigate the impact of these attacks. This includes designing robust multi-sensor fusion techniques to cross-verify sensor inputs, improving anomaly detection methods for early identification of adversarial behavior, and developing adaptive systems capable of learning and responding to emerging attack patterns over time.

\section{Conclusion} 
\label{conclusion}

In conclusion, this survey has explored the various adversarial attacks targeting LiDAR-based perception systems in autonomous vehicles and the corresponding defenses designed to mitigate these threats. We examined how these attacks exploit the vulnerabilities inherent in both sensor-specific and multi-sensor fusion-based models, leading to false positives and negatives that can severely compromise navigation and safety. From sensor spoofing to physical adversarial objects, these attack methods demonstrate the complex challenges faced in ensuring the reliability and robustness of perception systems.

Despite advances in defense mechanisms, such as model-agnostic and model-based strategies, current solutions are not foolproof and often fail to address the full range of potential attack scenarios, particularly in real-world environments. The transferability of attacks across different models and scenes, as well as the need for robust defenses that can function effectively in dynamic conditions, remains a critical area for future research. Moreover, enhancing the resilience of perception systems through multi-sensor fusion holds promise, but it requires systematic evaluation and refinement.

\section*{Acknowledgment}
This work was partially funded by the NYUAD Center for Cyber Security (CCS), funded by Tamkeen under the NYUAD Research Institute Award G1104 and the NYUAD Center for Interacting Urban Networks (CITIES), funded by Tamkeen under the NYUAD Research Institute Award CG001.
\bibliographystyle{unsrt}
\bibliography{bib.bib}

\EOD

\end{document}